\documentclass[10pt,journal,compsoc]{IEEEtran}
\ifCLASSOPTIONcompsoc
  \usepackage[nocompress]{cite}
\else
  \usepackage{cite}
\fi
\ifCLASSINFOpdf
   \usepackage[pdftex]{graphicx}
   \graphicspath{{figures}}
   \graphicspath{{people}}
   \DeclareGraphicsExtensions{.pdf,.jpeg,.png}
\else
\fi
\usepackage{amsmath}
\ifCLASSOPTIONcompsoc
 \usepackage[caption=false,font=footnotesize,labelfont=sf,textfont=sf]{subfig}
\else
 \usepackage[caption=false,font=footnotesize]{subfig}
\fi

\usepackage{microtype}                 
\PassOptionsToPackage{warn}{textcomp}  
\usepackage{textcomp}                  
\usepackage{mathptmx}                  
\usepackage{times}                     
\usepackage{cite}                      
\usepackage{tabu}                      
\usepackage{booktabs}                  
\usepackage{amsfonts}
\usepackage{xcolor}
\usepackage{multirow}
\usepackage{hyperref}
\usepackage[normalem]{ulem}
\usepackage{capt-of}
\usepackage{courier}  
\usepackage{fancyhdr}

\definecolor{gold}{RGB}{255,215,0}
\definecolor{silver}{RGB}{192,192,192}
\definecolor{bronze}{RGB}{191,173,111}

\newcolumntype{C}{S[table-format=1.3]}
\newcolumntype{U}[2]{S[table-format={#1}.{#2}]}

\newcommand{\bg}{{b_{g\vphantom{a}}}}
\newcommand{\ba}{{b_{\vphantom{g}a}}}

\newcommand\scalemath[2]{\scalebox{#1}{\mbox{\ensuremath{\displaystyle #2}}}}

\newcommand{\rdvio}{{RD-VIO}}
\newcommand{\sfvio}{{SF-VIO}}
\newcommand{\swvio}{{Baseline-VIO}}
\newcommand{\mh}{{\ttfamily MH}}
\newcommand{\mhI}{{\ttfamily MH\_01\_easy}}
\newcommand{\mhII}{{\ttfamily MH\_02\_easy}}
\newcommand{\mhIII}{{\ttfamily MH\_03\_medium}}
\newcommand{\mhIV}{{\ttfamily MH\_04\_difficult}}
\newcommand{\mhV}{{\ttfamily MH\_05\_difficult}}
\newcommand{\vIOI}{{\ttfamily V1\_01\_easy}}
\newcommand{\vIOII}{{\ttfamily V1\_02\_medium}}

\newcommand{\vIIOI}{{\ttfamily V2\_01\_easy}}
\newcommand{\vIIOII}{{\ttfamily V2\_02\_medium}}

\newif\ifhighlight
\highlightfalse 

\newcommand{\myhighlight}[1]{%
    \ifhighlight
        \textcolor{blue}{#1}%
    \else
        #1%
    \fi
}

\definecolor{gold}{RGB}{221, 196, 65}
\definecolor{silver}{RGB}{215, 215, 215}
\definecolor{bronze}{RGB}{126, 66, 5}

\usepackage{fancyhdr}  
  
\begin{document}

%
\title{RD-VIO: Robust Visual-Inertial Odometry for Mobile Augmented Reality in Dynamic Environments}

\author{Jinyu Li\IEEEauthorrefmark{1}, 
        Xiaokun Pan\IEEEauthorrefmark{1}, 
        Gan Huang,
        Ziyang Zhang, 
        Nan Wang, 
        Hujun Bao,
        Guofeng Zhang\IEEEauthorrefmark{2}

\IEEEcompsocitemizethanks{
\IEEEcompsocthanksitem  Jinyu Li, Xiaokun Pan, Gan Huang, Ziyang Zhang, Hujun Bao, Guofeng Zhang are with the State Key Lab of CAD\&CG, Zhejiang University, Hangzhou 310058,
China.\protect\\
E-mails: mail@jinyu.li, \{xkpan, huanggan, zhangzion, baohujun, zhangguofeng\}@zju.edu.cn
\IEEEcompsocthanksitem Nan Wang is with SenseTime Research, Hangzhou 311215,  China.\protect\\
E-mail: wangnan@sensetime.com

\IEEEcompsocthanksitem Digital Object Identifier 10.1109/TVCG.2024.3353263
}
}

\IEEEtitleabstractindextext{%

\begin{center} \setcounter{figure}{0}
  \includegraphics[width=0.9\linewidth]{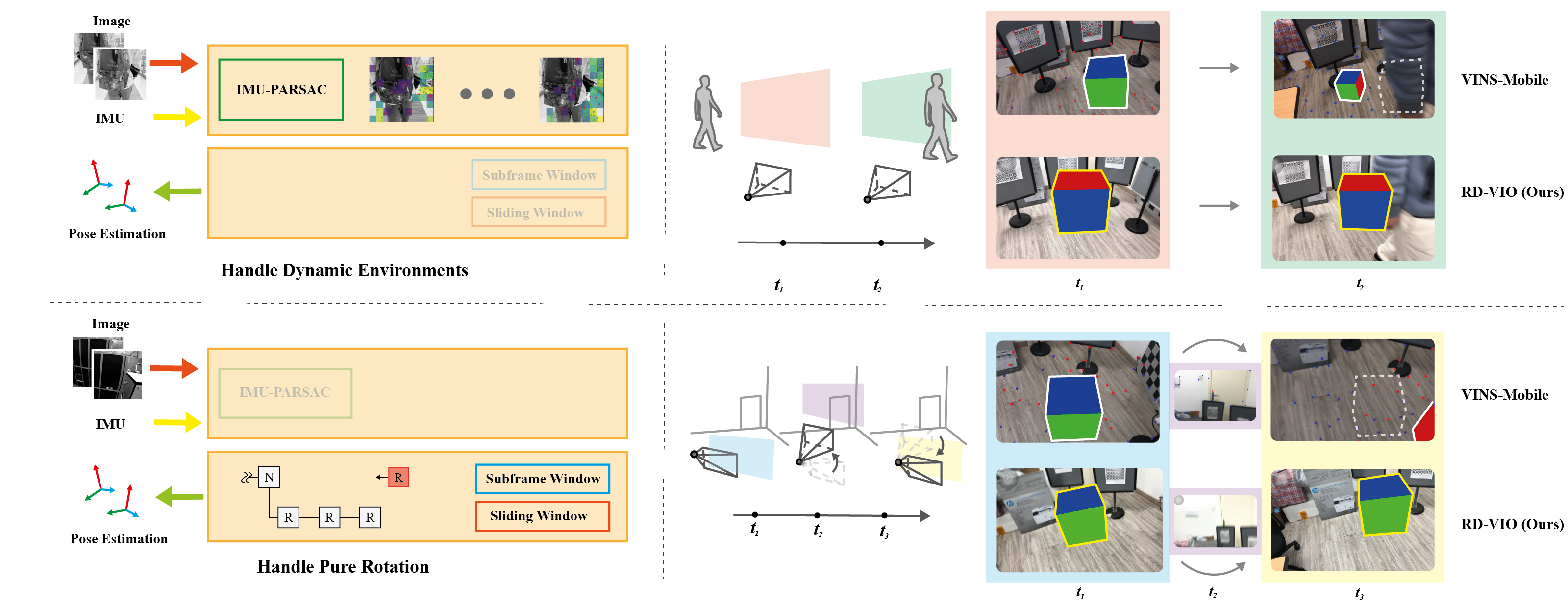}
  \captionof{figure}{The proposed RD-VIO can robustly work in dynamic scenes with pure rotation motions, and outperforms some other SOTA VIO/VI-SLAM systems such as VINS-Mobile.
  }
  \vspace{0.3cm}
  \label{fig:teaser}
\end{center}


\begin{abstract}
It is typically challenging for visual or visual-inertial odometry systems to handle the problems of dynamic scenes and pure rotation. In this work, we design a novel visual-inertial odometry (VIO) system called RD-VIO to handle both of these two problems. Firstly, we propose an IMU-PARSAC algorithm which can robustly detect and match keypoints in a two-stage process. In the first state, landmarks are matched with new keypoints using visual and IMU measurements. We collect statistical information from the matching and then guide the intra-keypoint matching in the second stage. Secondly, to handle the problem of pure rotation, we detect the motion type and adapt the deferred-triangulation technique during the data-association process. We make the pure-rotational frames into the special subframes. When solving the visual-inertial bundle adjustment, they provide additional constraints to the pure-rotational motion. We evaluate the proposed VIO system on public datasets and online comparison. Experiments show the proposed RD-VIO has obvious advantages over other methods in dynamic environments. The source code is available at: 
\href{https://github.com/openxrlab/xrslam}{{\fontfamily{pcr}\selectfont https://github.com/openxrlab/xrslam}}.
\end{abstract}

}

\maketitle

\renewcommand{\thefootnote}{\fnsymbol{footnote}}
\footnotetext[1]{Equal contribution}
\footnotetext[2]{Corresponding author}

\IEEEdisplaynontitleabstractindextext

%
\IEEEpeerreviewmaketitle

\thispagestyle{fancy}

\IEEEraisesectionheading{\section{Introduction}\label{sec:introduction}}

\renewcommand{\thefootnote}{\arabic{footnote}}


\IEEEPARstart{V}{}isual-inertial odometry~(VIO) systems are crucial to many applications from virtual reality~(VR), augmented reality~(AR), drones, autonomous driving, and robotics.
Most of the visual-inertial SLAM systems rely on VIO technique for sensor fusion and state estimation. 
Especially, a robust and light-weight VIO is very crucial for performing augmented reality on a mobile device. A bunch of algorithms have been developed to improve the accuracy and speed of VIO.

Compared to visual odometry, VIO systems can fuse visual and inertial information to achieve better robustness in complex environments, such as textureless scenes with dynamic objects. However, if there are large moving objects and degenerated motion~(like long-time stopping or pure-rotation), traditional VIO systems still easily encounter robustness problems. Although a few methods~\cite{bescos2018dynaslam}\cite{bescos2021dynaslam}\cite{yu2018ds}\cite{liu2022rgb} have been proposed to use semantic information to improve the tracking robustness in dynamic environments, the computational complexity is still a big problem to achieve real-time performance on a mobile device.

In this paper, we focus on robustifying a VIO system from two aspects: better moving keypoint removal and robust pure-rotation handling, while still keeping the system lightweight. To recognize moving keypoints, we propose a novel algorithm IMU-PARSAC which detects and matches keypoints in a two-stage process.
In the first stage, known landmarks are matched with new keypoints using both visual and IMU measurements. We collect error-statistics from the matching results, which then guide the intra-keypoint matching in the second stage.
To handle pure-rotation, we detect the motion type for the incoming image frames.
We adapt the deferred-triangulation technique during the data-association process, where we postpone the triangulation for landmarks under pure-rotational situations.
The pure-rotational motion information is honored in our modified design of sliding-window in a way that it always keeps keyframes with sufficient translations.
We make pure-rotational frames into special subframes.
When solving the visual-inertial bundle adjustment, they provide additional constraints to the pure-rotational motion.

As shown in Figure ~\ref{fig:teaser}, the proposed VIO system RD-VIO can accommodate pure-rotational motions and large moving objects, which would easily lead to divergence on many other VIO/VI-SLAM systems, such as VINS-Mobile~\cite{qin2017vins}.
We test the proposed system and compared it with many state-of-the-art VIO systems in public datasets.
The experimental results show that our proposed system not only produces accurate tracking results but also does so in a more robust manner.

The major contributions of this paper are as follows:
\begin{itemize}
\item A novel IMU-PARSAC algorithm is proposed to detect and remove moving ourliers in dynamic scenes, which can obviously improve the tracking robustness.
\item A novel subframes strategy in the sliding window is proposed to efficiently reduce drift under pure rotational motion.
\item The source code of the whole system is released to benefit the community, including core VIO algorithm and an iOS project for mobile AR application.
\end{itemize}

\section{Related Works}
For precise and dependable tracking in mobile augmented reality (AR) applications, especially within dynamic environments, it is imperative to devise robust visual-inertial odometry systems. Yet, conventional methods like VIO and SLAM sometimes falter in the face of intricate motions and ever-changing scenarios, leading to tracking discrepancies. Consequently, scholars are delving into innovative strategies, including deep learning-infused VIO and SLAM, along with adaptive SLAM techniques, to better navigate these hurdles and yield more exact tracking outcomes.
\subsection{VIO and SLAM} 
Visual odometry and SLAM systems are popular in recent vision studies. \cite{yousif2015overview} provides a useful review for visual-only SLAM systems.
However, vision-only systems cannot recover the metric scale of the scene.
In visual-inertial odometry~(VIO), the inertial measurements are fused with visual measurements to recover the metric scale.
According to their fusion technique, VIO systems can be roughly divided into filter-based and optimization-based.

 \textbf{Filter-based VIO} MSCKF~\cite{mourikis-icra-2007} \cite{li2012improving}  are early VIO systems based on Kalman filtering. The state vector of its filtering consists of a fixed number of frame poses.
Landmarks and their observations are processed and marginalized for each update phase.
So the amount of computation is bounded.
ROVIO~\cite{bloesch-ijrr-2017} is another filtering-based system that used photometric errors for visual observation.
So the data association is integrated with the filter estimation process.
It also demonstrates the use of bearing vectors in the parametrization of landmarks.
R-VIO\cite{huai2022robocentric} is a novel robocentric VIO algorithm. 
Different from the standard world-centric algorithms, R-VIO provides more accurate estimates of relative motion in relation to a moving local frame, and gradually updates the global pose.

OpenVINS\cite{geneva2020openvins} is a newly open-source platform using MSCKF filter. The modular design makes it flexible to use and easy to expand. Open source datasets evaluation show its high precision and robustness.  

In order to further improve the performance of VIO, some newly VIO systems like\cite{bao2022robust} use pre-built high-precision maps to improve accuracy greatly. And some others like RNIN-VIO\cite{chen2021rnin} take advantage of the neural network of IMU navigation to improve robustness.

\textbf{Optimization-based VIO }OKVIS~\cite{leutenegger-ijrr-2015} is an optimization-based system.
It works in a sliding window fashion by adding new keyframes into the optimization and marginalizing old keyframes.
The marginalization of the old frames linearizes old observations into priors terms and adds the prior into the optimization.
VINS-Mono~\cite{qin-tro-2018} and VINS-Fusion~\cite{qin2019a} \cite{qin2019b} \cite{qin-tro-2018} are recent VI-SLAM systems.
The frontend also uses keyframe-based bundle adjustment with a sliding window.
The loop-closure in the backend can help cancel accumulated errors, thus achieve better precision.
VI-ORB-SLAM~\cite{murartal-ral-2017} is a loosely-coupled VI-SLAM system, meaning that its inertial measurements are not fused immediately with visual observations.
For each new frame, visual observations are processed first, using the traditional VSLAM approach.
Then the result is aligned with inertial measurements to get the metric result.
In its recent evolution, ORB-SLAM3~\cite{campos2021orb} demonstrated a tightly-coupled system that produces astonishingly accurate results.
However ORB-SLAM3 solves the full SLAM problem, meaning that early poses still get optimized by using later observations.
This is simply impossible for real-time applications.
VI-DSO\cite{von2018direct} and DM-VIO\cite{stumberg22dmvio} are VIO extensions of original DSO, which increase robustness and have true scale.

\textbf{Deep learning based SLAM }Recently, deep learning based methods are widely adopted into SLAM systems, and achieved amazing results. DROID-SLAM\cite{droid_slam} uses a dense and accurate optical flow RAFT\cite{teed2020raft} as measurement, and builds an end-to-end network to perform bundle adjustment of pose and structure. PVO\cite{Ye2023PVO} has a further performance improvement by coupling panoramic semantic segmentation. NICE-SLAM\cite{zhu2022nice}, Vox-Surf\cite{li2022vox}, Vox-Fusion \cite{yang2022vox} and the latest ESLAM\cite{johari-et-al-2023}, Co-SLAM\cite{wang2023coslam} are typical neural implicit SLAM methods. With the powerful representation ability of neural implicit expression, these methods also show good performances both of tracking and mapping. But all of these methods depend on a large computation cost of CPU or GPU, which limits the application for AR.

\subsection{SLAM in Dynamic Environments} 
Robustness in dynamic environments is also a research hotspot in the SLAM field. Both traditional methods, such as RDSLAM~\cite{tan2013robust} based on the assumption of dynamic object distribution, and deep learning-based methods like DynaSLAM~\cite{bescos2018dynaslam}, DynaSLAM II~\cite{bescos2021dynaslam}, and DynaFusion~\cite{owada2008dynafusion}, have conducted in-depth research on this issue.  In VIO/VI-SLAM systems, due to the independence of IMU measurement from the external environment, they have a certain level of robustness in dynamic scenes compared to pure visual SLAM systems, resulting in relatively less research. Despite many successful VIO/VI-SLAM systems, bad visual cues can still damage the tracking quality. Since VIO systems are using static landmarks, keypoints from moving objects can have adverse effects. Hence in highly dynamic extreme environments, even using an IMU cannot prevent the system from being affected by dynamic objects. 

Typically, VIO systems rely on methods like the traditional robust estimator RANSAC~\cite{fischler-cacm-1981}. However, when moving keypoints \myhighlight{dominate the view}, these RANSAC systems usually do not work well. RDSLAM~\cite{tan2013robust} proposed an alternative method of evaluating model hypothesis. In their PARSAC algorithm, they exploits the locality of moving objects. Keypoints on these objects usually get clustered in a region so the model with mostly scattered keypoints will be elected as the background model. PARSAC works well for small-sized moving objects. However if the moving object is big enough to hijack the scene, the locality heruisic will fail. To get robust background model, some approaches rely on structure regularities like planes, nevertheless these approaches are primarily limited to particular scenarios. DynaVINS~\cite{song2022dynavins} introduced a robust bundle adjustment capable of discarding dynamic outliers by utilizing pose priors derived from IMU preintegration. We also use the strong prior of IMU measurement to efficiently estimate dynamic observation in a different way.
Thanks to the rapid development of deep learning, some systems such as Mask-SLAM~\cite{kaneko2018maskslam}, DS-SLAM~\cite{yu2018ds}, Dynamic-VINS~\cite{owada2008dynafusion}, SOF-SLAM~\cite{cui2019sofslam}, Mask-Fusion~\cite{runz2018maskfusion} use semantic segmentations to aid moving object handling. However, these methods are still very heavy-weight, which imposes another limitation on VIO applications.

\subsection{SLAM under Degenerated Conditions} Degenerated movements can also be problematic, and people have noticed this problem in prior works.
When the camera undergoes only negligible translation, there are few observations of the landmark depths, making it difficult to accurately estimate the camera's position and orientation. To address this issue, some previous works have proposed different approaches.
DT-SLAM~\cite{herrera-3dv-2014} attempted to avoid triangulating new landmarks in order to prevent initialization with erroneous depths. Instead, the system relied on existing landmarks and their known positions to estimate camera motion.
Another approach, described in \cite{pirchheim-ismar-2013}, involves tracking rotation-only cameras with panoramas, resulting in a system with mixed geometry models. This allows the system to handle cases where the camera is not translating, as well as cases where it is.
For the inertial measurements, stopped cases can be exploitable.

In IMU-only tasks, such as pedestrian navigation, one can detect full stops in the foot movement and use it for suppressing accumulation error.
This technique is commonly known as the zero-velocity update~(ZUPT)~\cite{nilsson-ipin-2014}.
The idea is to constrain the integration because we know the total translation should be zero when stopped.
LARVIO~\cite{qiu2020lightweight}, a recent VIO system, employed the idea of ZUPT to improve performance. By modeling ZUPT as an elegant closed-form measurement update, it has successfully achieved a trade-off between computational efficiency and localization precision

There are also works that aim to address the degradation problem by utilizing geometric features in the environment. When there are a large number of similar point features in the scene, point features are prone to degradation, leading to increased uncertainty in estimation or even complete failure of estimation. PL-VINS\cite{fu2020pl} and PL-VIO\cite{he2018pl} leverage a visual-inertial bundle adjustment to minimize separate reprojection errors for point features and line features in the visual measurement residuals. In addition, some works rely on structure of planes like RP-VIO\cite{ram2021rp} and PVIO\cite{jinyu-prcv-2019}, that uses planar features to increase robustness and accuracy in dynamic environments.

\section{Approach}

\begin{figure}[ht]
    \centering
    \includegraphics[width=\linewidth]{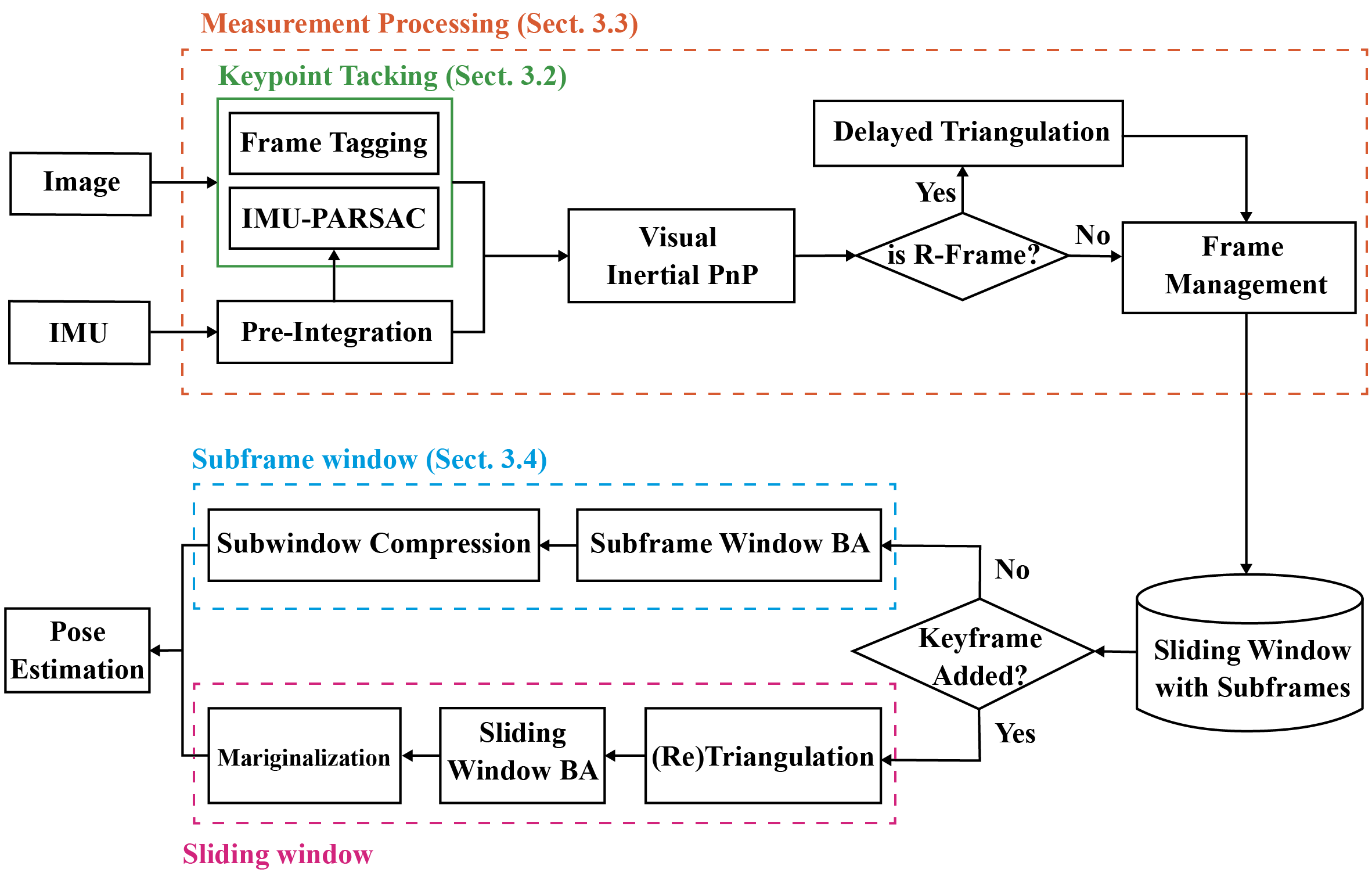}
    \caption{The pipeline of RD-VIO}
    \label{fig:pipeline}
\end{figure}

We begin with a baseline VIO system, which is based on PVIO~\cite{jinyu-prcv-2019} but without using planar prior. The baseline VIO will also be used for comparison. The pipeline of our system is as shown in Figure \ref{fig:pipeline}.
On the top of this system, we made our modifications.
We detect pure-rotations and triangulate landmarks properly, and then organize pure-rotational frames into subframes and optimize the poses accordingly.


\subsection{Sliding-Window VIO}
Our system followed a sliding-window approach.
So we first introduce a baseline VIO system with sliding window optimization (\swvio{}), and define most of the notations.
The \swvio{} system works by keeping a recent number of keyframes in a window, running bundle adjustment to fuse visual and inertial measurements, and marginalizing out stale frames, as if a multi-frame window sliding along the time.
\subsubsection{Sliding-Window Optimization}
During the tracking, we keep a fixed number of recent keyframes and the landmarks observed in these frames.
The states of these keyframes and landmarks will be refined with a visual-inertial bundle adjustment.
The state vector of a frame is parametrized as $s_i = [ p_i, q_i, v_i, \bg_i, \ba_i ]$ --- the position, orientation, velocity, gyroscope bias and accelerometer bias correspondingly.
Quaternion representation is used for $q_i$.
We use inverse-depth parametrization~\cite{civera-tro-2008} for landmarks.
The state of a landmark $x_k$ consists only a 1-dimensional inverse depth $d_k$.
For the sake of simplicity, we assume a constant camera intrinsic matrix $K$, and an ideal configuration between camera and IMU where their relative rotation and translation can be ignored (the extrinsic parameter between them is equal to identity).
For a landmark $x_k$, we denote its position on frame $i$ as $u_{ik}$.
This landmark is associated with a reference frame $r_k$ and a reference keypoint position $u_{{r_k}k}$.
So the 3D point corresponding to this landmark can be represented by $x_k = \frac{1}{d_k}C(q_{r_k}){K^{-1}\bar{u}_{r_kk}}/{\|K^{-1}\bar{u}_{r_kk}\|} + p_{r_k}$.
Here $C(q)$ denotes the rotation matrix of a quaternion $q$, and $\bar{u}$ denotes the homogeneous vector $\bar{u} = \begin{pmatrix}
    u \\ 1
\end{pmatrix}$.
When $x_k$ is observed in another frame $i \neq r_k$, let $\Pi$ be the homogeneous projection, we get the following reprojection error:
\begin{equation}
\scalemath{0.9}{
    E_{\mathrm{reproj}(i, k)} = \|\Pi [ K C^\top(q_i)(x_k - p_i) ] - u_{ik}\|^2.
    }
\end{equation}
To process IMU measurements,  we employ the method used in \cite{jinyu-prcv-2019} to add IMU cost term:
\begin{equation}
\scalemath{0.9}{
    \begin{aligned}
        E_{\mathrm{motion}(i,j)} =& D^2(q_j, \hat{q_j}) \\
        &+ \|v_j - \hat{v_j}\|^2 + \|p_j - \hat{p_j}\|^2 \\
        &+ \|\bg_j - \hat{\bg_j}\|^2 + \|\ba_j - \hat{\ba_j}\|^2.
    \end{aligned}
    }
\end{equation}
Here $\hat{q_j}, \hat{v_j}, \hat{p_j}, \hat{\bg_j}, \hat{\ba_j}$ are the measurements of motion state at IMU frame j by IMU pre-integration. 
$D(q_1, q_2) = \|\text{Log} (q_2^{-1} \cdot q_1)\|$ is the distance between two nearby quaternions $q_1$ and $q_2$ by recognizing their difference as a small purturbation in the underlying Lie-algebra.
All the norms here must be derived using Mahalanobis distances, with the corresponding covariances.
We skip the maths for computing covariances here, but a reader can refer to~\cite{forster2017manifold} for details.

The final bundle adjustment will solve for the states that minimize the total of all the reprojection errors as well as the motion measurement errors:
\begin{equation}\label{eq:bundle-adjustment}
    \scalemath{0.9}{
        \underset{\{s_i\}, \{d_k\}}{\arg\min} \sum_i \sum_k E_{\mathrm{reproj}(i, k)} + \sum_i E_{\mathrm{motion}(i, i+1)} + E_{\mathrm{marg}}.
    }
\end{equation}
$E_{\mathrm{reproj}}$ and $E_{\mathrm{motion}}$ are the cost terms of vision and IMU, respectively. $E_{\mathrm{marg}}$ is a prior term which comes from marginalization. Similar to~\cite{jinyu-prcv-2019}, we marginalize out the states of an old keyframe as soon as this keyframe is out of sliding window to bound the computational complexity.

\subsubsection{Initialization}

The initialization of the VIO includes the pursuit of the gravity vector, the solution to the global scale, and the determination for the initial states.
First, a sequence of initial frames are selected, and we do a visual-only SfM with these frames.
The result gives the relative pose of these frames up to some arbitrary scale.
Then the IMU measurements are aligned with the SfM results.
From the alignment, we can solve for gravity vector and the initial scale using the method introduced in~\cite{qin-iros-2017}.
Finally, a complete bundle adjustment~\eqref{eq:bundle-adjustment} is used for finding the best initial states.

\subsubsection{Keypoint Tracking}

We detect and track keypoints using KLT as described in ~\cite{gftt1994shi}.
If a keypoint already has an associated landmark, we predict its landing position on the next frame by projecting the landmark onto this new frame.
And we use this position as the initial position for KLT tracking.
The pose for the new frame, which has not been solved yet, is extrapolated by integrating IMU measurements since the last solved frame.
To get rid of outlier matches, we estimate an essential matrix and a homography matrix using RANSAC~\cite{fischler-cacm-1981}.
Instead of choosing between two results, like in ORB-SLAM~\cite{murartal-tro-2015}, we estimate the homography matrix after estimating the essential matrix.
So the second (homography) RANSAC can take advantage of the matches obtained from the previous (essential) one.
The essential matrix RANSAC is using a tight error threshold, which aimed to enforce binocular geometry relations.
Meanwhile, the homography RANSAC uses a much larger error threshold, in the sense that for relatively small movements, movements of keypoints can be loosely described by a homography.
This two-pass RANSAC solely relies on visual information, and can suffer from motion ambiguities.
Therefore, we propose an IMU-PARSAC algorithm to make improvements, which will be introduced in the next section.

After keypoint tracking, a new frame will be registered with the sliding window.
Let $s_i$ be the new frame's states, $\{x_k\}$ be the landmarks tracked in this frame.
We solve the following visual-inertial PnP to get an intial estimation to $s_i$:
\begin{equation}\label{eq:pnp}
    \underset{s_i}{\arg\min} \sum_k E_{\mathrm{reproj}(i, k)} + E_{\mathrm{motion}(i-1, i)}.
\end{equation}

A typical sliding-window based system, like OKVIS~\cite{leutenegger-ijrr-2015} or VINS-Mono~\cite{qin-tro-2018}, will conditionally mark the new frame as a keyframe.
If it is a keyframe, it will be optimized with a bundle adjustment like~\eqref{eq:bundle-adjustment}.
It will be appended to the back of the sliding-window, and the oldest frame will be marginalized out.
Otherwise, it will be quickly purged to save computation cost.
In the following sections, we introduce how we modify the keypoint tracking to reduce moving object matches and how we handle the new frame and the sliding-window to fight the low-translation problem.

\subsection{Outliers Detection and Removal}

\begin{figure}
    \centering
    \includegraphics[width=\columnwidth]{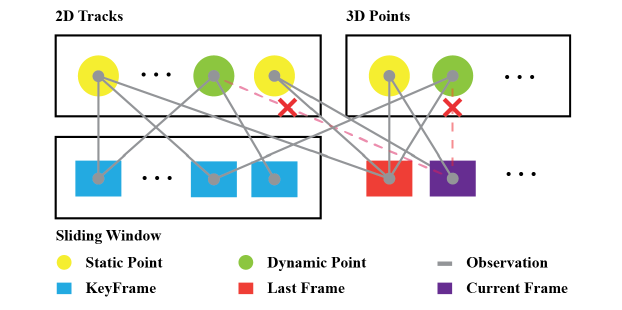}
    \caption{Moving outlier detection and removal strategy: In the mandatory 3D-2D stages, the current frame obtains initial matches of 2D observations and 3D points based on optical flow tracking with the last frame. After the IMU-PARSAC algorithm, most outliers are filtered out. In the optional 2D-2D stage, the current frame and the key frames in the sliding window are matched frame by frame using the original PARSAC algorithm. The remaining dynamic outliers are removed through this multi-view cross-validation approach.}
    \label{fig:outlier_removing}
\end{figure}

We introduce the IMU-PARSAC algorithm, leveraging IMU information to differentiate between moving elements and static backgrounds. This differentiation enhances the robustness of VIO tracking.
Our dynamic outlier removal approach unfolds in two phases: an essential 3D-2D matching phase (IMU-PARSAC) and an optional 2D-2D matching phase as shown in Figure~\ref{fig:outlier_removing}. In the initial phase, we align static 3D landmarks from the map to the 2D keypoints of the newly captured image. IMU preintegration predicts the current pose, guiding the 3D-2D matching process. If landmarks are scant, new ones are derived from 2D-2D matches. After this, we collect error statistics from the 3D-2D phase, formulating dynamic thresholds for 2D-2D PARSAC. This strategy counters the variable errors stemming from moving objects.
At its heart, our methodology seeks to weave IMU measurements into a robust parameter estimation algorithm framework, and synthesis capitalizes on the synergistic benefits of both the camera and IMU.

\subsubsection{3D-2D Matching Stage}

Upon the arrival of a new frame~$i$, we apply \eqref{eq:pnp} for the initial $s_i$.
So we need to match the 3D landmarks with the 2D keypoints on frame~$i$.
We are assuming that these landmarks were static at the time of triangulation.
And outlier matches are due to false correspondences or objects start moving.
Vision-only RANSAC algorithms can easily overlook the unwanted correspondences from the moving objects.
Therefore we propose to use the pose prediction from integrating the IMU measurements.
Our IMU-PARSAC algorithm followed the iterative scheme of the typical RANSAC~\cite{fischler-cacm-1981} algorithm.
For each iteration, after sampling a hypothesis it solves the model and creates the consensus correspondence set.
Let $q_\mathrm{VIS}, p_\mathrm{VIS}$ be the model fit from visual correspondences.
At the same time, we have $q_\mathrm{IMU}, p_\mathrm{IMU}$ predicted by applying pre-integration from frame~$i-1$ to frame~$i$.
We can obtain the visual-consensus set $S_\mathrm{VIS}$ and motion-consensus set $S_\mathrm{IMU}$ as:
\begin{equation}
    \scalemath{0.95}{
    \begin{aligned}
        S_\mathrm{VIS} &= \{(x_k, u_{ik})\,|\,E_{\mathrm{reproj}(i, k)}(q_\mathrm{VIS}, p_\mathrm{VIS}) \leq \epsilon_\mathrm{VIS}^2\}, \\
        S_\mathrm{IMU} &= \{(x_k, u_{ik})\,|\,E_{\mathrm{reproj}(i, k)}(q_\mathrm{IMU}, p_\mathrm{IMU}) \leq \epsilon_\mathrm{IMU}^2\}.
    \end{aligned}
    }
\end{equation}
We then take their intersection $S_\mathrm{VI} = S_\mathrm{VIS} \cap S_\mathrm{IMU}$ as the final consensus set for this iteration.
Note that for each frame ${i}$, $S_\mathrm{IMU}$ is not changing between IMU-PARSAC iterations.
Therefore we can pre-compute $S_\mathrm{IMU}$ and \myhighlight{look for its ``best'' subset}.

Now, in terms of the ``best'', we no-longer use the number of correspondences $|S_{VI}|$ or the total re-projection error $\sum E_{\mathrm{reproj}(i, k)}$.
Following the prior-based adaptive RANSAC algorithm proposed by \cite{tan2013robust}, we evaluate the distribution of the inlier keypoints with the weighted covariance: 
\begin{equation}
    \scalemath{0.85}{
    \mathrm{Cov}(S_\mathrm{VI}) = \frac{\sum \lambda_k}{(\sum \lambda_k)^2 - \sum \lambda_k^2} \sum_k \lambda_k (u_k - \mathrm{B}[u_k])(u_k - \mathrm{B}[u_k])^T.
    }
\end{equation}
 $\mathrm{B}[u_k]$ represent the center coordinate of the bin which 2D observation $u_k$ located in, and $\lambda_k$ is the inlier ratio of each bin as the confidence weight, but it is different from the PARSAC that the time prior is considered in our method.

In some highly dynamic scenes, it may not be sufficient to identify static landmarks based solely on the information from a single frame. In such cases, historical information is needed. This is because static landmarks in the scene can be triangulated and tracked stably for a longer period of time. To account for this, we take into consideration the observation time prior to the hypothesis evaluation step. Specifically, we collect the continuous observation time of each landmark and use the average time of all observations in each bin to measure the motion state of that bin. This is similar to the original PARSAC algorithm, which uses the proportion of inliers in each bin as the confidence weight. These weights determine the influence of each bin on the current model evaluation. To improve the accuracy of the weights, we add a prior time factor to the redefinition of the weight.

\begin{equation}
\scalemath{0.9}{
    w_{i}^{t} = 1 - p^{0.1t},
}
\end{equation}
\begin{equation}
\scalemath{0.9}{
    \lambda_{i}^{'} = w_{i}^{t}\cdot \lambda_{i}.
}
\label{eq:prior-time-weight}
\end{equation}
In Equation~(\ref{eq:prior-time-weight}), $t$ represents the average observation time of all landmarks in $b_{i}$. We define the observation time as the number of continuous frames that can observe the landmark. To measure the dynamic degree of the scene, we introduce a dynamic coefficient $p \in [0,1]$. This coefficient can be adjusted to fit a particular scene. A larger value of $t$ indicates that the landmark is more likely to be static and should be given more weight than newly generated landmarks when evaluating hypotheses.

Then we can define the quality of the consensus set as:
\begin{equation}
\scalemath{0.9}{
    a(S_\mathrm{VI}) = \sum \lambda_k \cdot \sqrt{\mathrm{det}(\mathrm{Cov}(S_\mathrm{VI}))}.
}
\end{equation}
The $S_\mathrm{VI}$ with highest $a(S_\mathrm{VI})$ will be selected as our final inlier set.
Geometrically, it corresponds to the inlier set whose keypoints are most scattered on the image.

This combination of $S_\mathrm{VI}$ and $a(S_\mathrm{VI})$ has several implications:
a) The difference between the vision-based pose estimation and IMU-based prediction impacts the number of correspondences in $S_\mathrm{VI}$.
b) Even when there is a tie in correspondence count, the quality criteria selects the most spreaded subset of keypoints. But when a huge rigid body is moving in the view, the use of IMU prediction will help avoid creating correspondence on this body. Otherwise, the system could fixate on this moving object, thinking it is the static background, and following adrift.
c) In case when visual estimation and IMU prediction are having a large difference, it suggests that either there are a lot of moving outliers moving around or the IMU prediction is inaccurate. There will be insufficient correspondences in $S_\mathrm{VI}$, leading us to the 2D-2D stage. 

\subsubsection{2D-2D Matching Stage}

In our sliding-window strategy, failing to track a landmark will lead to this landmark being marginalized.
So we need to populate new landmarks to keep them at a sufficient number.
The 2D-2D stage is used for this purpose.
In this stage, we used the original PARSAC algorithm from~\cite{tan2013robust}.
However, near-degraded cases can still be challenging.
Inside the PARSAC algorithm, we rely on the epipolar geometry of the matches.
Since outliers are determined based on the epipolar distances, when keypoints are moving along the direction of the epipolar lines, these moving-object keypoints will be indistinguishable from the static background.
Luckily, in the real world, noises in the motion will lead to errors in the epipolar distance.
With the help of IMU data, we can identify moving keypoints in the 3D-2D stage without worrying about the motion characteristics.
So we can train an epipolar distance threshold from the 3D-2D matches.
And use the trained value for the 2D-2D stage thresholding.

Let $N_I$ and $N_O$ be the number of inliers and outliers from a 3D-2D stage.
We sort the result matches according to the epipolar distances.
Let $\{d^+_0, d^+_1, \dots, d^+_{N_I}\}$ be the epipolar distances of the inlier matches in non-descreasing order, $\{d^-_0, d^-_1, \dots, d^-_{N_O}\}$ be the distances of the outlier matches correspondingly, and $\lambda^+, \lambda^- \in [0,100]$ be the prechosen percentile value.
We update the 2D-2D stage threshold $\epsilon_\mathrm{2D}$ as:
\begin{equation}
\scalemath{0.9}{
    \epsilon_\mathrm{2D} = \frac{d^+_{\lfloor \lambda^+ \cdot N_I \rfloor} + d^-_{\lfloor \lambda^- \cdot N_O \rfloor}}{2}.
}
\end{equation}
$\epsilon_\mathrm{2D}$ will be updated whenever $N_I + N_O \geq N_P$, $N_P$ is a predefined total number.

We do not triangulate the 2D-2D inlier matches immediately.
Instead, we trace the historical matches.
If a keypoint is observed $L$ times previously, and be marked as an inlier for $L^+$ times whereas $L^+ \geq \delta L$, we regard it as a static keypoint and label it for triangulation.
However, its triangulation can still be postponed, as we will introduce next. 

\subsection{Pure-Rotation Detection and Delayed Triangulation}

\begin{figure}
    \centering
    \includegraphics[width=\linewidth]{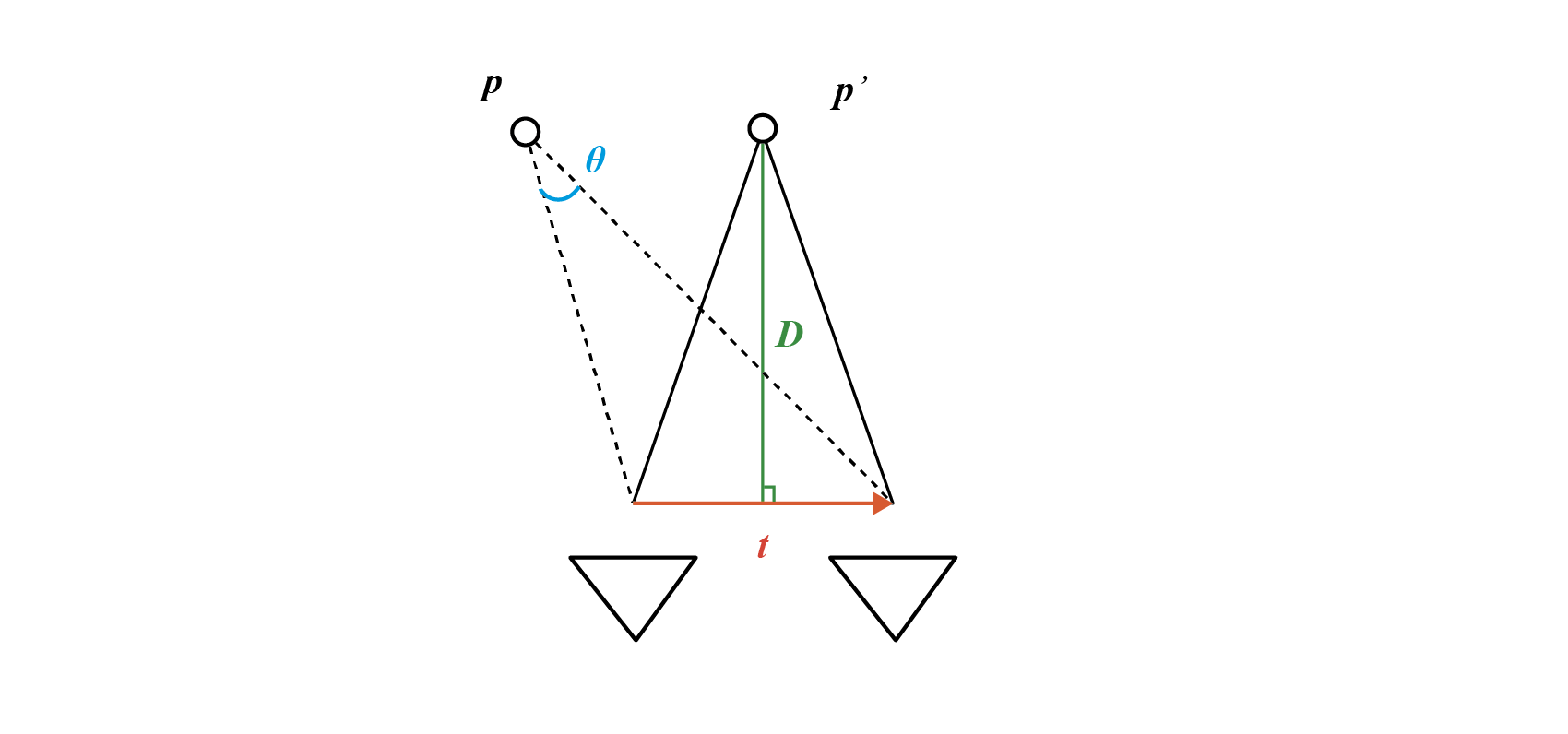}
    \caption{Geometry illustration of our angle based pure-rotation detection. The maximum $\theta$ is realized when two rays-of-observation and the translation vector $t$ forms an isosceles triangle.}
    \label{fig:pure-rotation-detection}
\end{figure}

\begin{figure}
    \centering
    \begin{tabular}{cc}
        \includegraphics[width=0.21\textwidth]{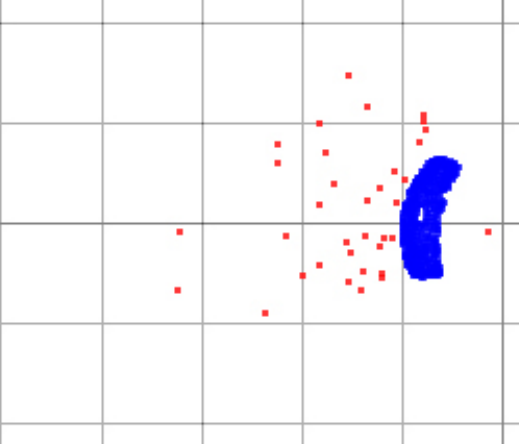} &
        \includegraphics[width=0.21\textwidth]{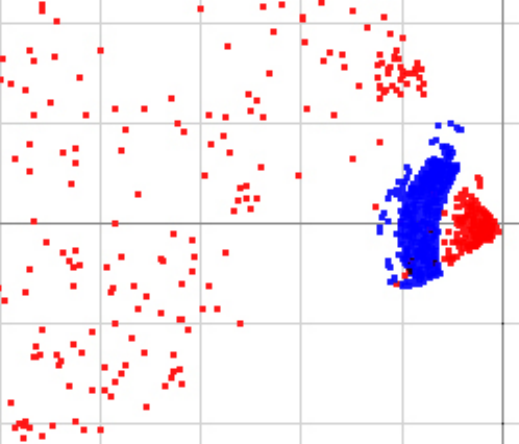} \\
        \scalebox{0.85}{\swvio{} + DT} & \scalebox{0.85}{\sfvio{}}
    \end{tabular}
    \caption{Example for point cloud from tracking when camera is stopped. Blue points are DT landmarks. They are casted into points with a fake 1m depth for visualization. And we can see normal landmarks (in red) are scarse in \swvio{}, because their depths are diverging. With DT, more keypoints can be tracked. \sfvio{} on the otherhand, can keep the depths stable.}
    \label{fig:delayed-triangulation}
\end{figure}



Due to the large noise of the IMU sensors from consumer-level phones, we design a vision-based method to detect pure-rotation. We add a third RANSAC pass in feature tracking. The 3rd one is solving for a rotation matrix from the matches. Suppose there is a translation $t$ between the latest two frames. The two frames are observing a common landmark $p$. Let $D$ be the distance from $p$ to the line where $t$ belongs. From the geometry relation, as shown in Figure~\ref{fig:pure-rotation-detection}, the angle between the two observations must satisfy
\begin{equation}
\scalemath{0.9}{
    \theta \leq 2 \arctan \frac{\|t\|}{2D}.
    }
\end{equation}
Therefore, when $\|t\| \ll D$, i.e. the translation is insignificant comparinig the the depth, the two observations will be nearly in-parallel.
Conversely, if all the keypoints' motion can be well described by a common rotation matrix, then the frame is likely performing a pure rotation.
Based on this observation, we compare the angles between the bearing vectors of the matching keypoints.
Let $R_{ij}$ be the rotation matrix from frame $j$ to frame $i$, $\{(u_{ik}, u_{jk})\}$ be the positions of the matching keypoint-pairs,
the maximum angle of observations is computed as:
\begin{equation}
\scalemath{0.9}{
    \theta_{\max} = \max \{ \langle \bar{u}_{ik}, R_{ij} \bar{u}_{jk} \rangle \}.
}
\end{equation}
If $\theta_{\max} \leq \theta_{\mathrm{rot}}$, a pre-defined threshold, we tag the latest frame as ``pure-rotational-frame'', or R-frame for short, otherwise, it is a ``normal-frame'' or an N-frame.

If one frame is an R-frame, it lacks depth observation of new landmarks.
When new keypoints are detected from this frame, we choose to triangulate them into landmarks partially.
We only record the originating frame $r_k$ and the location $u_{r_kk}$ of a new landmark $k$, delaying the estimation of its depth $d_k$.
Upon getting sufficient observations of the depths, we re-estimate and update these landmarks.
This delayed-triangulation~(DT) strategy is inspired by~\cite{herrera-3dv-2014}.
Figure~\ref{fig:delayed-triangulation} shows how DT can help tracking more keypoints even for \swvio{}.
But since \swvio{} cannot fully leverage the DT information in its bundle adjustment,
in experiments, \swvio{} is not using any DT.

It worth mentioning that an R-frame is visually incapable of providing depth observations.
We can extract some early position estimation, for example, through a VI-PnP.
However, in some situations, like when the translation is small or the surrounding space is huge, depth observations can still be insufficient, leading to significant triangulation error.
Therefore, our vision-based method fits well in this visual-inertial fusion context.

\subsection{Sliding Window with Subframes}

Like we introduced before, we cannot afford to fill the sliding window with R-frames, nor can we discard R-frames as they must be kept for continuous estimation of IMU biases.
\myhighlight{We introduce a \textbf{subframe} mechanism in our system, which allows a keyframe to carry a set of subframes, as illustrated in the lower branch of each case in Figure ~\ref{fig:frame-management} (the N-frame which they attached to known as keyframe). The system utilizes this subframe strategy to handle long sequences of R-frames. The design rationale behind the subframe mechanism will be gradually explained in this section.}

\subsubsection{Frame Management}

\begin{figure}
    \centering
    \includegraphics[width=1.0\linewidth]{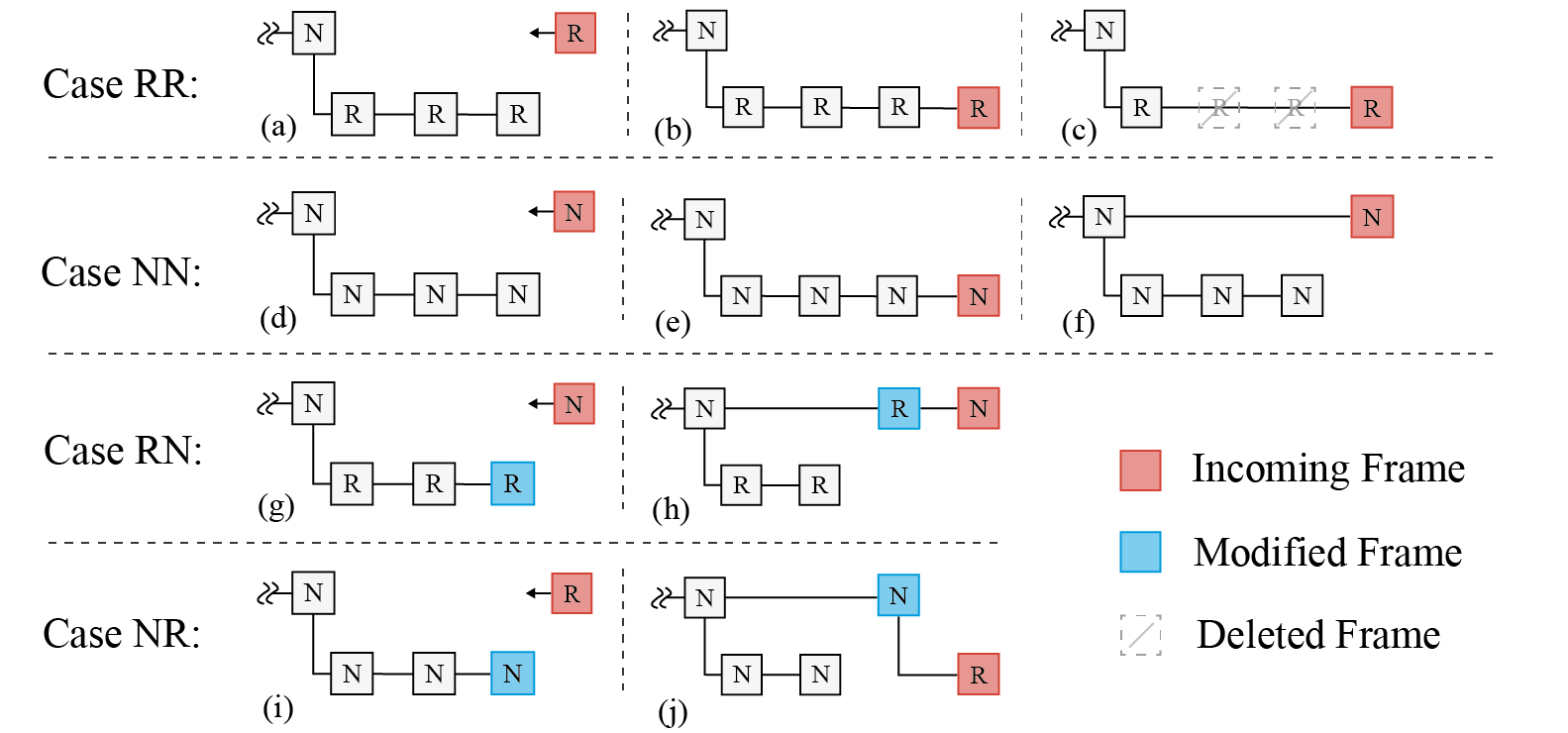}
    \caption{Frame management rules for adding new frames at the tail of the sliding window. (a,d,g,i) are the initial cases while (b,e,f,h,j) are the results after new frames added. (c) demostrates R-frame compression introduced in section \ref{sec:r-frame-ba}. }
    \label{fig:frame-management}
\end{figure}

The strategy of frame management is the key to our sliding window structure.
We follow several rules to add the new frame into the sliding window.
The following two properties are kept when adding a new frame:
\begin{itemize}
    \item The last keyframe in the sliding window is always an N-frame,
    \item There will never be N-frames and R-frames mixed in the same subframe window.
\end{itemize}

Based on the type of frames in the latest subframe window and the type of the new frame,
there will be $2\times 2 = 4$ cases.
Figure~\ref{fig:frame-management} illustrates all the frame management rules.
In the following text, if the subframes are R-frames and the new frame is an N-frame, this is called case RN for simplicity.

Assume there are already subframes in the latest keyframe.
For the RR case, we will append the new frame to the end of the subframes.
When there are consecutive new R-frame comes, a series of RR-case will happen so that the last subframe will be extended with R-frames.
This is important since we can have a chain of IMU preintegrations between these R-frames.
The positions of these R-frames will be constrained in a ZUPT-like manner, while the orientations can be aligned with either 3D landmarks or delay-triangulated bearing vectors.
The detailed optimization for these R-subframes will be introduced later.
And we will introduce our remedy to avoid excessively long R-subframes when the device is not translating for a while.
In a word, the subframe of R-frames will allow us better modeling the transitioning IMU biases during a pure-rotational motion.

For case NN, since we do not lack any depth observations, the new N-frame can always be used as a keyframe.
Yet, we want to save some computations, so we followed the idea used in VINS-Mono.
When the number of N-subframes exceeds a predefined size $N_s$, the new N-frame will be added as a keyframe, as shown in Figure~\ref{fig:frame-management}(f).
Otherwise, it will be added as another subframe, as shown in Figure~\ref{fig:frame-management}(e).

For case RN, we first make the last R-subframe into a keyframe, then add the new N-frame as a keyframe.
For case NR, we make the last N-subframe a keyframe, then add the new R-frame as a subframe to this keyframe.
In this way, the two properties of the sliding window will always be ensured.

Besides the above cases, when the last keyframe has no subframes, the new frame will be added as a subframe regardless of its type.
Also, when the number of tracked keypoints is below some threshold $N_t$, the new frame will be added as an N-keyframe anyway.

\subsubsection{Bundle Adjustment}

With the modified sliding window, the bundle adjustment will make use of the two properties mentioned above.
When there is no new keyframe, we don't run full BA.
Instead, we optimize the states in the last subframe window for a quick update.
If the last subframe window contains N-frames, we have sufficient translation, hence sufficient depth observations.
The BA we used here is the same as~\eqref{eq:bundle-adjustment}, but with keyframes and landmarks observed in these keyframes fixed.
In this case, only new landmarks observed in the last subframe window, as well as the states of these subframes, are refined.
The result of this BA registers the new landmarks with the old landmarks, but we don't have to solve the full sliding window because there will be at most $N_s$ subframes in an $N$-type subframe window.

On the other hand, if the last subframe is filled with R-frames, we will deal with the chain of preintegrations for better IMU bias estimation.
We give up estimating depths since R-frames means insufficient translation.
Hence, we regularize the orientation of the subframes with the bearing vectors from delayed triangulation.
The orientation error of subframe $i$ when observing a delay-triangulated landmark $k$ can be written as:
\begin{equation}
    \scalemath{0.8}{
        E_{\mathrm{rot}(i,k)} = \left\| C^\top(q_i)C(q_{r_k})\frac{K^{-1}\bar{u}_{r_kk}}{\|K^{-1}\bar{u}_{r_kk}\|} - \frac{K^{-1}\bar{u}_{ik}}{\|K^{-1}\bar{u}_{ik}\|} \right\|^2.
    }
\end{equation}
We also fix the keyframe pose and old landmarks in the optimization.
So they can be used to constrain the poses of these subframes.
Since the relative translation should be tiny, a ZUPT-like regularizer can also be used.
For example $E_{\mathrm{ZUPT}(i)} = \| p_i - p_{i-1} \|^2$.
This is trivial but can be helpful when the number of landmarks is small.

The overall optimization for this R-type subframe window will be:
\begin{equation}
    \scalemath{0.9}{
        \begin{aligned}
        \underset{\{s_i\}}{\arg\min} & \sum_i \sum_{k, d_k \neq 0} E_{\mathrm{reproj}(i, k)} + \sum_i \sum_{k, d_k = 0} E_{\mathrm{rot}(i, k)} \\
        & + \sum_i E_{\mathrm{motion}(i, i+1)} + \sum_i E_{\mathrm{ZUPT}(i)}.
        \end{aligned}
    }
    \label{eq:r-bundle-adjustment}
\end{equation}
Here, $d_k = 0$ designates the landmarks that are delay-triangulated (i.e. $d_k$ hasn't been computed yet).
Since all the keyframes are fixed, $E_{\mathrm{marg}}$ will be constant, hence ignored in~\eqref{eq:r-bundle-adjustment}.
$E_{\mathrm{reproj}}$, $E_{\mathrm{rot}}$ and $E_{\mathrm{ZUPT}}$ all contributes to the stabilization of the subframe poses.
Allowing the IMU-biases be better optimized through minimizing $E_{\mathrm{motion}}$.
The regularization from $E_{\mathrm{rot}}$ and $E_{\mathrm{ZUPT}}$ may temporarily lead to larger localization error in favor of robustness.
However, experiments show that the compromise on accuracy is small.
But we gain better stability during degenerate movements.
In the example shown in Figure~\ref{fig:delayed-triangulation}, more landmarks are preserved with the help of R-frames.

Unlike the N-type subframe window, there is no number limit for R-frames in a subframe window.
If there are too many of them, solving~\eqref{eq:r-bundle-adjustment} will be slow.
As a remedy, the subframe window is compressed when the total number of R-frames exceeds $N_r$.
We evenly choose $\lceil 1/3 \rceil$ of the R-frames and concatenate the preintegrations in between.
Figure~\ref{fig:frame-management}(c) illustrates this operation when $N_r = 3$\label{sec:r-frame-ba}.
Since~\eqref{eq:r-bundle-adjustment} can produce good estimation to intermediate poses and IMU-biases.
We won't lose too much accuracy in this compression.
Otherwise, the estimation of the IMU-biases will be more easily trapped into a local minimum in later optimizations.

When there are new keyframes added into the sliding window, we perform a full BA on all the keyframes.
This is similar to~\eqref{eq:bundle-adjustment}.
Except that for keyframes carrying R-type subframes, the chain of preintegrations is used instead.
After BA, if the sliding window has more than $N_w$ keyframes, we marginalize the keyframes from the oldest to the newest until there are $N_w$ keyframe left.
Since the subframes are not participating in this keyframe BA, they are removed when we marginalize a keyframe.

For the convenience of writing and ablation study, we name SF-VIO as the version with only subframe strategy based on our baseline VIO.
\section{Experiments}

To evaluate the effectiveness of our proposed method and the robustness of the VIO system, we conducted a series of experiments. Regarding the dynamic outlier removal strategy, we qualitatively compared and analyzed IMU-PARSAC algorithm with other algorithms. To address the problem of system state estimation degradation caused by pure rotation, we investigated the detection performance of pure rotation and the system stability when the camera is stationary. Finally, we quantitatively compared our method with current state-of-the-art VIO/VI-SLAM algorithms on publicly available datasets.

We evaluated our method and other SOTA systems on two public datasets.

\begin{itemize}
\setlength{\itemindent}{0em}
\item 
The EuRoC~\cite{Burri25012016} dataset is a benchmark dataset for VIO and SLAM algorithms. It includes high-quality data captured by a micro aerial vehicle (MAV) equipped with a stereo camera and a synchronized IMU, covering various indoor scenarios. The dataset provides ground truth poses obtained from a motion-capture system, which enables the evaluation of the accuracy of the estimated trajectories. 

\item 
ADVIO~\cite{cortes2018advio} (Advanced Visual-Inertial Odometry) dataset is an open benchmark dataset designed for evaluating visual-inertia algorithms. It contains diverse real-world scenes, including different indoor/outdoor environments, lighting conditions, and dynamic object interferences, which can be used to evaluate the robustness, accuracy, and real-time performance of various algorithms. 

\end{itemize}


\subsection{Outliers Removal}

\begin{figure}
    \centering
    \includegraphics[width=\columnwidth, trim={0cm 0cm 4cm 0cm}, clip]{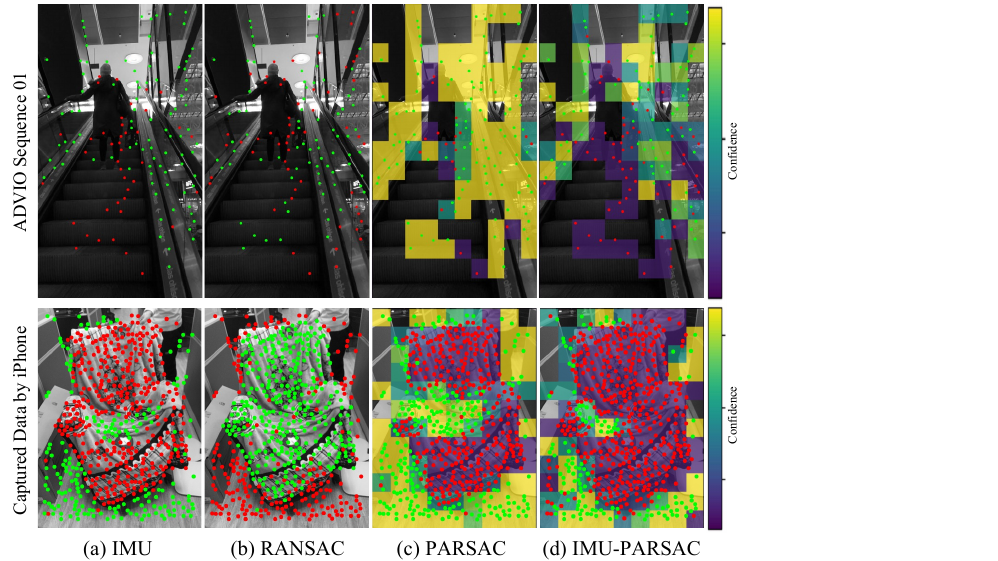}
    \caption{Qualitatively comparison between several outlier removal methods: (a) using IMU pre-integration predicted pose to identify outliers (b) traditional robust etimator RANSAC (c) dynamic object distribution prior estimator PARSAC (d) our proposed IMU-PARSAC algorithm. We visualize the 2D observations and color them as green for inliers and red for outliers based on the inlier mask. Additionally, PARSAC and IMU-PARSAC are visualized their corresponding bins based on the confidence.} 
    \label{fig:imu-parsac-compare}
\end{figure}

We conducted qualitative and quantitative evaluations of IMU-PARSAC on both handcrafted scenes and public datasets ADVIO. Our handcrafted scenes consisted of static backgrounds and moving objects in the foreground, with some objects occasionally occluding a significant portion of the field of view to test the ability of IMU-PARSAC. Figure~\ref{fig:imu-parsac-compare} compares several outlier removal schemes across two different scenarios, including traditional robust estimator RANSAC, dynamic object distribution prior estimator PARSAC, and our proposed IMU-PARSAC. We also compared the ability to eliminate outliers in visual observations using IMU pre-integration predicted poses. We performed PnP geometric estimation on the 2D points visible in the current frame and the 3D points in the map to identify whether the observed 2D points correspond to moving objects. 

The top row of Figure~\ref{fig:imu-parsac-compare} is from Sequence 01 of ADVIO~\cite{cortes2018advio}, where the operator walks in a shopping center and rides an escalator. When on the escalator, the operator is not walking and the device maintains a gaze on the steps of the escalator. To accurately estimate the current state of the device, the 2D visual observations should all come from static structures in the environment. We visualized the results of several outlier removal methods and found that neither IMU pre-integration predicted poses nor RANSAC could completely detect out the dynamic outlier (escalator). The bottom row shows a dynamic scene captured with an iPhone, where a chair in the middle of the image is moved forward, and all 2D observation points falling on the chair should be considered outliers. 
In this scenario, the method of IMU pre-integration can only remove a portion of the outliers, while RANSAC completely estimated an incorrect geometric model which led to a “hijacking” situation, as RANSAC can only consider the geometric model with the greatest number of fitted points as the final result.
At the same time PARSAC algorithm makes assumptions about the inliers distribution, it is difficult to apply to more general scenarios, and it is challenging to completely remove dynamic points. Our IMU-PARSAC algorithm, utilizing the motion-consensus property of IMU, can find the correct inliers even in highly dynamic scenes.

The quantitative evaluation of the dynamic object removal strategy is in Section 4.4.2.

\subsection{Detection of Pure Rotation}
\begin{figure}[!htb]
    \includegraphics[width=0.45\textwidth]{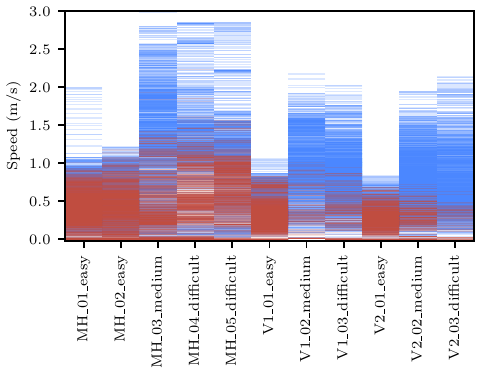}
    \caption{Heatmap of R and N-frames.Red for R-Frames and blue for N-Frames}
    \label{fig:r-frame-hist}
\end{figure}

To carefully examine the pure-rotation detection and stabilization effect, we rely on the high-quality ground-truths from the EuRoC datasets.
We computed the motion speed from the ground-truth data and plot the speed curve. For each R-frame detected, we add a red line indicating its time point.
For all the sequences, there are long stopped periods.
And our method was able to mark almost all the frames in these periods as R-frames.
In fact, only a few frames in \mhI{}, \mhII{}, and \mhV{} are the outliers.
They are due to moving objects appeared in the background.
Besides stopped periods, we can see many speed local-minimums successfully detected as R-frames.

The scene appeared in \mh{} sequences is large, and the overall motion speed in \vIOI{} and \vIIOI{} is slow.
Therefore we can see sparsely marked R-frames in a lot of the local-minimum points.
To further examine the speed range for our pure rotation detection method, \myhighlight{we draw the heatmaps of R-frames and N-frames for each sequence in Figure \ref{fig:r-frame-hist}}. R-frames are distributed in the lower part of the speed range. Due to the large scale of the scene in \mh{} sequences, the hot zone of R-frames reaches slightly higher speed ranges for these sequences.

\subsection{Stabilization Effect}

\begin{figure}
    \includegraphics[width=0.45\textwidth]{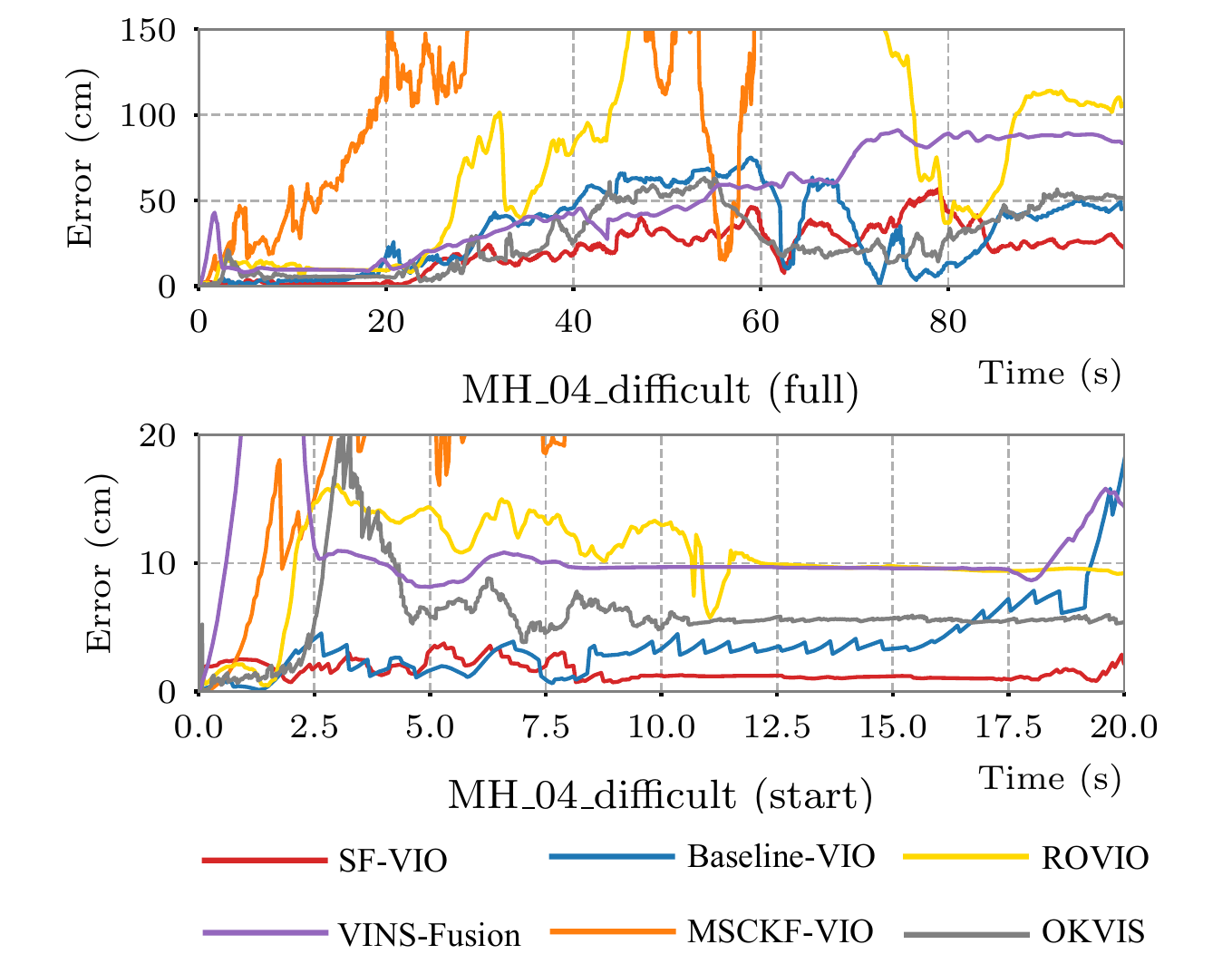}
    \caption{Positioning error curves for the first 20 seconds of the sequence \mhV{}.}
    \label{fig:position-error-curve}
\end{figure}

\begin{figure*}[!htb]
    \centering
    \includegraphics[width=0.95\textwidth]{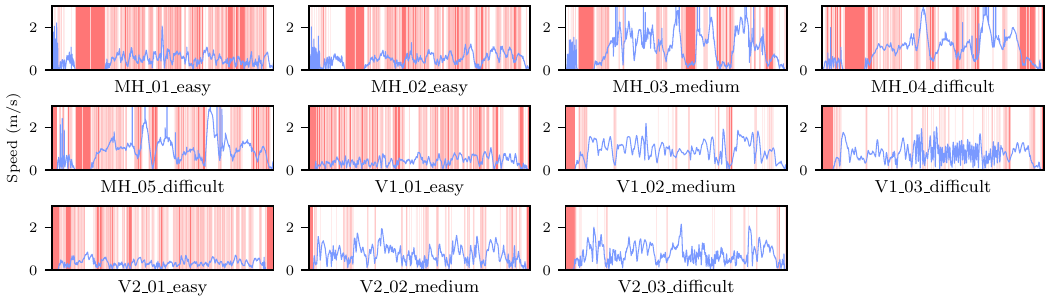}
    \caption{Speed plot for all the sequences with R-frames highlighted. Red lines denote the detected R-Frames and the speed curves are expressed in blue.}
    \label{fig:r-frame-plot}
\end{figure*}

The main goal of the subframes is the better handling of low-translation scenarios.
In the EuRoC dataset, all the sequences have a long period of stopping in the beginning.
By running an algorithm and examine the behavior at these stopped cases, we can see if it can handle them well.
To compare the results, we align the result trajectories at the starting point, then minimize their overall RMSEs by registering them with the best global rotation. This is done using the Umeyama algorithm~\cite{umeyama-tpami-1991}.
Then we can compare the poses with the ground-truth and plot the curve of positioning error.
\myhighlight{In Figure~\ref{fig:position-error-curve}, we compare the error curves of \sfvio{} and \swvio{} on sequence \mhV{}.}

The error curve for \swvio{} is jittering when the drone is stopped: its position slowly drifts, then suddenly being ``dragged'' back to a better location, then drifts again.
This drift is the consequence of missing depth observation -- the positions of the landmarks are becoming more and more inaccurate, and the error in IMU integration is accumulating.
Whenever there is a small purturbation in the translation, the keyframe is added into the sliding window. The bundle adjustment will quickly pick up the new depth observation and correct the pose of the new keyframe, resulting in the ``jittering'' result in its error curve.
Sometimes, the lack of proper depth observation causes \swvio{} produces jumping results, as can be seen around 10th to 11th seconds in sequence \mhV{}.
If the stopping continues, \swvio{} will be starved of reliable landmarks eventually.

On the other hand, \sfvio{} was able to accommodate the stopping situation.
As shown in in Figure~\ref{fig:r-frame-plot}, We have visualized the speed curves of the groundtruth trajectories for each sequence, along with the detection results of the R-Frames. We can tell that the frames during the stopped time are identified as R-frames, hence are handled in the subframe window.
The compression strategy was able to keep the subframe window short while keeping the IMU bias estimation stable.
In both sequences, the algorithm can keep the tracked position ``locked'' to its place.
Therefore, we can see a smooth and flat error curve during these periods.

\subsection{Qualitative Comparison}

\newcommand{\0}{\hphantom{0}}
\newcommand{\1}[1]{{\color{red}#1}}
\newcommand{\2}[1]{{\color{blue}#1}}

Besides the above evaluations, we also compute the RMSE of the localization error for quantitative comparison.
We compute the RMSE of each result using the tool {\ttfamily evo}\footnote{\url{https://github.com/MichaelGrupp/evo}}.
It registers the trajectory result with the ground-truth by seeking the RMSE-minimizing global rigid transform.
For each sequence, we run an algorithm 10 times, and take the average RMSE as the final result. 
We compared \sfvio{} with \swvio{}, VINS-Fusion, ORB-SLAM3, OKVIS, ROVIO, MSCKF and PVIO~\cite{jinyu-prcv-2019}, \textit{etc}.
Except for VINS-Fusion and ORB-SLAM3, all other systems are VIO systems based on optimization or Kalman filter.
VINS-Fusion is a complete VI-SLAM system which uses a sliding window based VIO frontend for tracking.
PVIO is a VIO system using plane (structural) heuristics for regularizing the sliding window.
Our system, on the other hand, uses motion heuristics.
In addition,  we use the third-party implementation of VIO based on MSCKF\footnote{\url{https://github.com/daniilidis-group/msckf_mono}} for comparison.

Although there has been a lot of work focus on the SLAM systems in dynamic scenes, most of the work is based on pure visual methods, such as RDSLAM~\cite{tan2013robust}, DynaSLAM~\cite{bescos2018dynaslam}, DynaSLAMII~\cite{bescos2021dynaslam}, or based on RGBD input, such as DynaFusion~\cite{owada2008dynafusion}. This is quite different from our VIO system, so we do not consider VSLAM solutions in the experiments. At the same time, as we emphasize the robustness of a lightweight VIO system on mobile devices in dynamic scenes, recent approaches based on dynamic object segmentation, such as Dynamic-VINS~\cite{liu2022rgb}, have high requirements for GPU computation and are difficult to achieve real-time performance even on PCs. Therefore, we do not compare our method with such approaches. DynaVINS~\cite{song2022dynavins} is the newest VIO system designed for handling dynamic scenes. We compared the robustness against our method in dynamic scenarios with it.


We evaluate two different versions of VINS-Fusion and ORB-SLAM3 with/without using loop closure and global bundle adjustment.
To ensure a fair comparison in terms of accuracy and real-time performance, we firstly report the online pose estimation results of VINS-Fusion and ORB-SLAM3 without loop closure optimization or final global optimization. 
For this assessment, we disabled the loop closure and global bundle adjustment modules in backend. We denote this version with ``VIO$^\ddagger$'' which represents the real-time estimated pose derived from the frontend.
Notably, the loop-closure component of ORB-SLAM3 is deeply coupled with the entire system, making it unfeasible to fully disable its loop-closure.
We also report the results of \textit{full} versions, which indicate the comprehensive use of all available information for pose optimization. We still use the real-time output camera poses of VINS-Fusion~(\textit{full}) for tracking accuracy evaluation, and found that the overall accuracy is slightly improved because of the loop closure optimization.
It should be noted that ORB-SLAM3 (\textit{full}) actually refines the whole camera trajectory by a post-processing, which indeed dramatically improves the RMSE but is meaningless for online applications (e.g. mobile AR applications).



\subsubsection{EuRoC Datasets}

\begin{table*}[tbh!]
    \caption{Tracking Accuracy (RMSE in meters) on the {EuRoC} Dataset. We highlight the top 3 results of each column in \colorbox{gold}{gold}, \colorbox{silver}{silver}, and \colorbox{bronze}{bronze}.}
    \centering
    \begin{tabular}{@{}l|cccccccccccc}
    \toprule
    Algorithm & MH-01 & MH-02 & MH-03 & MH-04 & MH-05 & V1-01 & V1-02 & V1-03 & V2-01 & V2-02 & V2-03 & AVG  \\ 
    \cmidrule{1-13}
    \rdvio{} & {\colorbox{gold}{\textbf{0.109}}} & {\colorbox{bronze}{\textbf{0.115}}} & {\colorbox{silver}{\textbf{0.141}}} & 0.247 & 0.267 & {\colorbox{bronze}{\textbf{0.060}}} & 0.091 & 0.168 & {\colorbox{silver}{\textbf{0.058}}} & 0.100 & {\colorbox{bronze}{\textbf{0.147}}} & {\colorbox{silver}{\textbf{0.136}}} \\
    \sfvio{} & {\colorbox{gold}{\textbf{0.109}}} & 0.147 & {\colorbox{gold}{\textbf{0.131}}}  & {\colorbox{silver}{\textbf{0.189}}} & {\colorbox{silver}{\textbf{0.240}}} & {\colorbox{gold}{\textbf{0.056}}} & 0.101 & {\colorbox{bronze}{\textbf{0.134}}}                   & {\colorbox{bronze}{\textbf{0.066}}} & 0.089 & {\colorbox{gold}{\textbf{0.122}}} & {\colorbox{gold}{\textbf{0.125}}} \\
    \swvio{} & 1.622 & 0.218 &  {\colorbox{bronze}{\textbf{0.154}}} & 4.313 &  {\colorbox{bronze}{\textbf{0.256}}} & 0.080 &  {\colorbox{silver}{\textbf{0.082}}} & 0.248 & 0.074 & {\colorbox{gold}{\textbf{0.073}}} & 0.289 & 0.674 \\ 
\cmidrule{1-13}
LARVIO & 0.132 & 0.137 & 0.168 & 0.237 & 0.314 & 0.083 &{\colorbox{gold}{\textbf{0.064}}}  & {\colorbox{silver}{\textbf{0.086}}} & 0.148 & {\colorbox{bronze}{\textbf{0.077}}} & 0.168 & {\colorbox{bronze}{\textbf{0.147}}} \\
Open-VINS & {\colorbox{bronze}{\textbf{0.111}}} & 0.287 & 0.181 & {\colorbox{gold}{\textbf{0.182}}}  & 0.365 & {\colorbox{silver}{\textbf{0.059}}} & {\colorbox{bronze}{\textbf{0.084}}} & {\colorbox{gold}{\textbf{0.075}}}  & 0.086 & {\colorbox{silver}{\textbf{0.074}}} &{\colorbox{silver}{\textbf{0.145}}}  & 0.150 \\
VI-DSO & 0.125 & {\colorbox{gold}{\textbf{0.072}}} & 0.285 & 0.343 & {\colorbox{gold}{\textbf{0.202}}} & 0.197 & 0.135 & 4.073 & 0.242 & 0.202 & 0.212 & 0.553 \\
OKVIS & 0.342 & 0.361 & 0.319 & 0.318 & 0.448 & 0.139 & 0.232 & 0.262 & 0.163 & 0.211 & 0.291 & 0.281 \\
MSCKF  & 0.734 & 0.909 & 0.376 & 1.676 & 0.995 & 0.520 & 0.567 & -     & 0.236 & -     & -     & 0.752 \\
PVIO & 0.129 & 0.210 & 0.162 & 0.286 & 0.341 & 0.079 & 0.093 & 0.155 &{\colorbox{gold}{\textbf{0.054}}} & 0.202 & 0.290 & 0.182 \\
DynaVINS & 0.308 & 0.152 & 1.789 & 2.264 & - & - & 0.365 & - & - & - & - & 0.976 \\
VINS-Fusion (VIO$^\ddagger$) & 0.149 & {\colorbox{silver}{\textbf{0.110}}} & 0.168 & {\colorbox{bronze}{\textbf{0.221}}}                   & 0.310 & 0.071 & 0.282 & 0.170 & 0.166 & 0.386 & 0.190 & 0.202 \\
ORB-SLAM3 (VIO$^\ddagger$) & 0.543 & 0.700 & 1.874 & 0.999 & 0.964 & 0.709 & 0.545 & 2.649 & 0.514 & 0.451 & 1.655 & 1.055 \\ 
\cmidrule{1-13}
VINS-Fusion (\textit{full}) & 0.178                  & 0.105                   & 0.143                   & 0.216                   & 0.362                   & 0.066                   & 0.287                   & 0.169                   & 0.131                   & 0.226                   & 0.173                   & 0.187                        \\
ORB-SLAM3 (\textit{full})   & 0.025                  & 0.048                   & 0.035                   & 0.088                   & 0.058                   & 0.043                   & 0.020                   & 0.031                   & 0.050                   & 0.017                   & 0.027                   & 0.040 \\
\bottomrule
\end{tabular}
    \label{tab:euroc}
\end{table*}

Table~\ref{tab:euroc} lists all the EuRoC RMSEs we gathered on these algorithms.
Comparing with \swvio{}, \sfvio{} showed significant improvements on many sequences.
Especially for sequence \mhI{} and \mhIV{}, thanks to the additional stabilization effect, the significant drifts are canceled.
We did carefully tuned \swvio{}.
And it turns out that even this baseline version can perform well on some of the sequences like \vIOII{} and \vIIOII{}.
However, the margin between \sfvio{} and \swvio{} on these sequences are small.
Even for \vIOII{}, \sfvio{} is only falling $1.9\mathrm{cm}$ behind.
Among all the VIO systems tested, we can see that \sfvio{} has top-tier accuracy with 6 of 11 sequences ranked within the top 3.
The subframe window came as a regularizer for small translation, and it did not compromises accuracy too much.

As there are almost no moving objects in the EuRoC, our proposed dynamic object removal strategy theoretically cannot improve the system's accuracy. On the other hand, any false detection of moving keypoints will lower the number of observations, resulting in a slight worse in RMSEs, such as in \mhIII{}, \mhIV{}, \mhV{}. However, we can see that \rdvio{} can achieve higher RMSEs compared to {} on some sequences, such as \mhII{} and \vIIOI{}. The reason is that although the proposed dynamic object removal strategy is not designed specifically for EuRoC, it still can remove the matches with relatively large errors, thereby improving the system's accuracy.


Furthermore, the complete VI-SLAM systems~(e.g., VINS-Fusion and ORB-SLAM3) generally combine VIO with loop closure and global bundle adjustment to eliminate the accumulated error. Nevertheless, \sfvio{} can provide a decent VIO frontend for these complete SLAM systems. Also, comparing with PVIO, which uses multi-plane heuristics, the compromise of our pure-rotation regularization is relatively small.

It is worth noting that for some of the compared methods, due to their poor initialization performance, a longer time was required for system initialization, which resulted in relatively shorter evaluated trajectory lengths. While this may have benefited the final RMSEs comparison, it is a disadvantage for the stability of a VIO/SLAM system. For example, OpenVINS requires 40 seconds for initialization on \mhI{} and 35 seconds on \mhII{}.

Meanwhile, to evaluate the system's operational efficiency, we also compared the running time on \vIOI{} with VINS-Mono. VINS-Mono is a sliding window optimization-based SLAM system that has been open-sourced. Our system has a similar structure to it, and after aligning the parameters, it is easy to compare the computational time of each part with VINS-Mono.
We measure the running time for each module of the system. We configured VINS-Mono with a sliding window size of 8 frames and deactivated its backend, ensuring a fair comparison between the two systems. Both VINS-Mono and RD-VIO were executed on a computer equipped with an Intel i7-7700 CPU @3.6GHz and 16GB of memory. The results for different modules are presented in Table~\ref{tab:time}.

\begin{table}[htbp]
\caption{Running time (ms) of different modules in VINS-Mono (front-end) and RD-VIO.}
\centering
\renewcommand{\arraystretch}{1.3}
\begin{tabular}{c|c c}
\hline
\multirow{2}{*}{Module} & \multirow{2}{*}{VINS-Mono} & \multirow{2}{*}{RD-VIO} \\
& & \\
\hline
Keypoint Tracking & 8.34 & 7.47 \\
Pre-Integration & 0.44 & 0.04 \\
Non-Keyframe PnP & 17.78 & 1.27 \\
Non-Keyframe Marg & 0.68 & - \\
IMU-PARSAC & - & 1.07 \\
Keyframe BA & 19.18 & 30.9 \\
Keyframe Marg & 32.91 & 3.99 \\
Keyframe Average & 60.87 & 42.4 \\
All frame Average & 44.72 & 18.38 \\
\hline
\end{tabular}

\label{tab:time}
\end{table}

For Non-Keyframe PnP, the full subframe window bundle adjustment (BA) is not executed. Instead, only the latest subframe's observations are used to quickly update the state. To maintain consistency, keyframes' states and observed landmarks remain fixed during subframe window BA. VINS-Mono conducts sliding window BA for every frame. In contrast, our method conducts subframe sliding window BA for non-keyframes, which significant reduces the time of non-keyframe PnP. As for Keyframe BA, when a new keyframe enters the sliding window, the entire main sliding window, including all its subframes, undergoes optimization. This process requires a longer computation time compared to VINS-Mono. Implementing the improved marginalization strategy outlined in PVIO\cite{jinyu-prcv-2019} allows for a more efficient approach, which allows us to effectively reduce the running time. Overall, the average computational time for all frames indicates that our system is more time-efficient than VINS-Mono.

\subsubsection{ADVIO Datasets}

\begin{table*}[tbh!]
\caption{Accuracy \& Completeness on the ADVIO Dataset. We highlight the top 2 results of each row in \colorbox{gold}{gold} and \colorbox{silver}{silver}. }
\centering
\begin{tabular}{c|ccccc|ccc}
\toprule
\multicolumn{1}{l|}{\multirow{2}{*}{sequence}} & \multicolumn{5}{c|}{RMSE} & \multicolumn{3}{c}{Comp.(\%)}                                                                                            \\
\multicolumn{1}{l|}{}                          & \multicolumn{1}{l}{~ ~SF-VIO~ ~} & \multicolumn{1}{l}{RD-VIO\textsuperscript{s1}} & \multicolumn{1}{l}{~RD-VIO} & \multicolumn{1}{l}{VINS-Fusion} & \multicolumn{1}{l|}{~ ~LARVIO~ ~} & \multicolumn{1}{l}{~RD-VIO} & \multicolumn{1}{l}{VINS-Fusion} & \multicolumn{1}{l}{~ ~LARVIO~ ~}  \\ 
\cmidrule{1-9}
01  & 2.177     & \colorbox{silver}{\textbf{1.956}}                            & \colorbox{gold}{\textbf{1.788}}                             & 2.339                           & 5.049                             & 97.9                                               & 59.6                            & 80.8                              \\
02                                             & \colorbox{gold}{\textbf{1.679}}         & 2.090                                              & \colorbox{silver}{\textbf{1.695}}                            & 1.914                           & 4.242                             & 96.8                                               & 68.2                            & 62.9                              \\
03                                             & 2.913                            & \colorbox{gold}{\textbf{2.270}}                             & 2.690                                              & \colorbox{silver}{\textbf{2.290}}         & 4.295                             & 98.7                                               & 70.1                            & 57.3                              \\
04                                             & -                                & -                                                  & \colorbox{gold}{\textbf{2.860}}                             & \colorbox{silver}{\textbf{3.350}}                           & -                                 & 89.9                                               & 71.2                            & -                                 \\
05                                             & 1.385                            & 1.366                                              &\colorbox{silver}{\textbf{1.263}}                            & \colorbox{gold}{\textbf{0.938}}          & 2.034                             & 95.1                                               & 67.7                            & 61.6                              \\
06                                             & \colorbox{silver}{\textbf{2.837}}          & 3.107                                              & \colorbox{gold}{\textbf{2.497}}                             & 11.005                          & 8.201                             & 97.7                                               & 70.0                            & 56.8                              \\
07                                             & \colorbox{silver}{\textbf{0.559}}          & 0.567                                              & \colorbox{gold}{\textbf{0.548}}                             & 0.912                           & 2.369                             & 87.7                                               & 82.2                            & 44.5                              \\
08                                             & 2.075                            & \colorbox{silver}{\textbf{2.009}}                            & 2.151                                              & \colorbox{gold}{\textbf{1.136}}          & 2.078                             & 90.6                                               & 64.5                            & 69.5                              \\
09                                             & \colorbox{gold}{\textbf{0.332}}           & 2.488                                              & 2.281                                              & \colorbox{silver}{\textbf{1.063}}         & 3.168                             & 94.7                                               & 69.1                            & 94.6                              \\
10                                             & 1.997                            & \colorbox{gold}{\textbf{1.700}}                             & 2.128                                              & \colorbox{silver}{\textbf{1.847}}         & 4.742                             & 99.2                                               & 73.9                            & 97.6                              \\
11                                             & \colorbox{silver}{\textbf{4.103}}          & 4.496                                              & \colorbox{gold}{\textbf{3.986}}                             & 18.760                          & 5.298                             & 99.6                                               & 70.4                            & 82.0                              \\
12                                             & 2.084                            & 2.032                                              & \colorbox{silver}{\textbf{1.951}}                            & -                               & \colorbox{gold}{\textbf{1.191}}            & 99.4                                               & -                               & 64.6                              \\
13                                             & 3.227                            & -                                                  &\colorbox{silver}{\textbf{2.899}}                            & -                               & \colorbox{gold}{\textbf{1.324}}            & 97.5                                               & -                               & 97.6                              \\
14                                             & \colorbox{gold}{\textbf{1.524}}           & -                                                  & \colorbox{silver}{\textbf{1.532}}                            & -                               & -                                 & 94.4                                               & -                               & -                                 \\
15                                             & \colorbox{silver}{\textbf{0.779}}          & \colorbox{gold}{\textbf{0.772}}                             & 0.780                                              & 0.944                           & 0.851                             & 94.2                                               & 68.9                            & 96.4                              \\
16                                             & \colorbox{silver}{\textbf{0.986}}          & \colorbox{gold}{\textbf{0.954}}                             & 0.991                                              & 1.289                           & 2.346                             & 98.0                                               & 68.1                            & 92.7                              \\
17                                             & 1.734                            & 1.862                                              & 1.657                                              & \colorbox{gold}{\textbf{1.235}}          & \colorbox{silver}{\textbf{1.569}}           & 99.9                                               & 70.0                            & 98.3                              \\
18                                             & 1.171                            & \colorbox{gold}{\textbf{1.057}}                             & \colorbox{silver}{\textbf{1.164}}                            & -                               & 3.436                             & 99.1                                               & -                               & 98.7                              \\
19                                             & 3.256                                & \colorbox{silver}{\textbf{2.740}}                                                  & 3.154                                                  & -                               & \colorbox{gold}{\textbf{2.010}}            & 59.4                                                  & -                               & 98.6                              \\
20                                             & -                                & \colorbox{gold}{\textbf{6.960}}                            & \colorbox{silver}{\textbf{7.013}}                            & 10.433                          & 16.441                            & 99.6                                               & 68.6                            & 98.2                              \\
21                                             & 8.962                            & \colorbox{gold}{\textbf{8.432}}                             & \colorbox{silver}{\textbf{8.534}}                            & 11.004                          & 13.142                            & 99.3                                               & 73.1                            & 97.9                              \\
22                                             & 4.686                            & \colorbox{gold}{\textbf{4.498}}                             & \colorbox{silver}{\textbf{4.548}}                            & -                               & 8.104                             & 99.8                                               & -                               & 98.7                              \\
23                                             & 6.631                            & \colorbox{silver}{\textbf{5.085}}                            & 6.486                                              &\colorbox{gold}{\textbf{4.668}}         & 9.389                             & 99.6                                               & 70.4                            & 98.0           \\        \bottomrule
\end{tabular}

\label{tab:advio}
\end{table*}

\myhighlight{As a challenging dataset in real-world settings, ADVIO offers 23 diverse scenarios, encompassing indoor and outdoor environments, varying lighting conditions, and dynamic elements such as pedestrians and vehicles.}
We found our proposed algorithms performed well on the ADVIO~\cite{cortes2018advio} datasets while most of the aforementioned algorithms didn't survive including current SOTA full VI-SLAM system ORB-SLAM3 and recent DynaVINS specialized in handling dynamic environment.
Beside our system, only VINS-Fusion and LARVIO were able to produce meaningful results.
In addition to accuracy comparison, we measured the trajectory completeness.
For each input image frame, if the visual tracking was able to give a meaningful pose output, it contributes to our completeness evaluation.
When a system is initializing, lost, or drift away by a large distance, the output is discarded, making the trajectory incomplete.

As the ADVIO dataset contains common dynamic scenes in some sequences, such as escalator, pedestrians, and trains, the results on this dataset can reflect the system's robustness to dynamic scenarios. In addition to reporting the results of full \rdvio{} system, we evaluated the effectiveness of our proposed two-stage dynamic object removal strategy by comparing it with \sfvio{}, which completely disables dynamic object removal strategy, and RD-VIO\textsuperscript{s1}, which only retains the IMU-PARSAC algorithm in the first stage. Since the completeness of trajectories depends on the system's initialization and termination time, and above three VIO systems have no difference in the initialization stage, most of the completeness of trajectories are the same. So we only report the trajectory completeness of the full RD-VIO system.

Table~\ref{tab:advio} lists the accuracy and completeness results from the ADVIO datasets.Compared to SF-VIO without dynamic object removal strategies, RD-VIO showed significantly better RMSEs on ADVIO dataset, and achieved the best accuracy on most of the sequences in both RD-VIO\textsuperscript{s1} and RD-VIO. Although RD-VIO\textsuperscript{s1} outperformed RD-VIO on some sequences, stage 2, achieved higher accuracy in some more complex dynamic scene sequences such as seq04, seq07, and seq09. And there is no significant difference on other sequences, even though the accuracy was slightly lower than stage 1.

While VINS-Fusion ranks as the second-most accurate algorithm, its result completeness significantly trails that of the two other VIO algorithms. A lower completeness, which indicates late initialization or tracking loss, results in shorter trajectories. These abbreviated trajectories can inherently lead to a higher RMSE. Hence, relying exclusively on RMSE as a measure might not provide a holistic view and could be deemed misleading.
Also, it failed on more sequences than LARVIO and \rdvio{}.
LARVIO could track most of the sequences and had a relatively good trajectory completeness thanks to its ZUPT scheme.
However, dynamic objects still negatively affect their tracking quality.
There are pedestrians and escalators in sequences 02-08.
In sequence 04, the operator rides multiple escalators consecutively.
LARVIO cannot handle these moving objects, therefore having a bad trajectory completeness or even failing to track.
The same situation happened for sequences 11 and 12 where subway trains are moving in the picture.
In comparison, \rdvio{} was able to recognize these moving bodies and robustify its tracking.
We can see \rdvio{} producing much more complete trajectories comparing with the other two algorithms.

\subsubsection{Online Comparison}
\begin{figure}
    \includegraphics[width=0.5\textwidth]{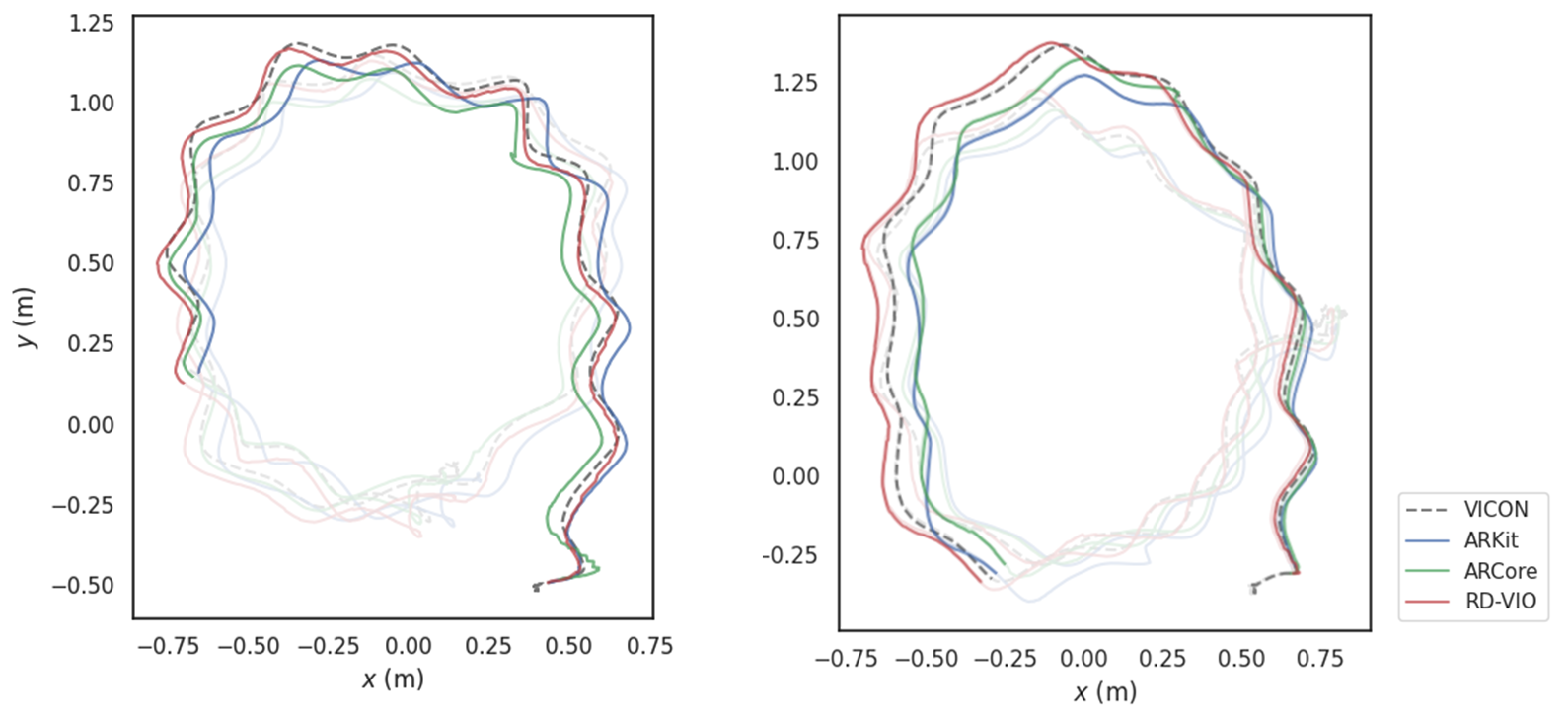}
    \caption{Trajectories of: VICON, ARCore, ARKit, RD-VIO on A3 (left) and A4 (right). To ensure clear visualization, the overlapping area is manually faded.}
    \label{fig:ARKit-Compare}
\end{figure}

In the previous section, various open-source algorithms were compared using pre-recorded data. However, since the two mainstream AR commercial software, ARKit\footnote{\url{https://developer.apple.com/documentation/arkit/}} and ARCore\footnote{\url{https://developers.google.com/ar/}}, can only directly access data from the device's camera and IMU, and do not support pre-recorded data, it is impossible to directly evaluate our system using the evaluation data from the previous section.  To facilitate a comparison between our work and ARKit/ARCore, we designed an online comparative experiment. The evaluation hardware configuration comprised two iPhone Xs and one Xiaomi Mi 8 smartphone. Three phones were tied together as close as possible on the same plane, facing the same direction. ARCore ran on the Xiaomi 8, ARKit ran on an iPhone X, and another iPhone X was used to record data for the execution of RD-VIO. Before the comparison, we performed an extrinsic calibration between the cameras of three smartphones and the VICON markers. We obtained the ground-truth camera trajectories with VICON. We simulated these scenarios using five cases (A0-A4). Case A0 features slow movement in a static scene. Cases A1 and A2 simulate rapid rotation, translation, and oscillation to test the tracking robustness under fast motion. Case A3 depicts pedestrians entering and exiting a room, creating a dynamic scene. Lastly, case A4 entailed manually occluding the camera for a certain period, which made tracking significantly more challenging. The experiment utilized the tracking accuracy and robustness\cite{jinyu2019survey} to evaluate the tracking quality of AR-based odometry and SLAM systems.

\begin{table}[h]
\caption{Comparison with ARKit \& ARCore. The best results of each row are highlighted in blod.}
\centering
\renewcommand{\arraystretch}{1.5}
\begin{tabular}{c|c|c c c}
\hline
Metrics      & Sequence & ARKit  & ARCore & RD-VIO \\ \hline
\multirow{5}{*}{APE (mm)}  & A0       & \textbf{15.643} & 16.625 & 22.385 \\ 
                           & A1       & 28.361 & \textbf{23.973} & 28.902 \\ 
                           & A2       & 22.576 & 31.576 & \textbf{20.931} \\ 
                           & A3       & 33.339 & 40.387 & \textbf{19.583} \\ 
                           & A4       & 42.615 & 34.492 & \textbf{23.049} \\ \hline
\multirow{5}{*}{Robustness} & A0       & \textbf{0.078}  & 0.083  & 0.112  \\ 
                           & A1       & 0.142  & \textbf{0.120}  & 0.145  \\ 
                           & A2       & 0.113  & 0.158  & \textbf{0.105}  \\ 
                           & A3       & 0.167  & 0.202  & \textbf{0.098}  \\ 
                           & A4       & 0.213  & 0.172  & \textbf{0.115}  \\ \hline
\end{tabular}
\label{tab:comparison}
\end{table}
Table~\ref{tab:comparison} shows the absolute position error (APE) of 3 algorithms in millimeters and their corresponding robustness values, where smaller values indicate better performance. Compared to ARKit and ARCore, our system registers slightly larger APE in static scenes with typical camera motion (e.g. A0). However, it performs on par with ARKit and ARCore in fast-paced scenes (e.g. cases A1 \& A2).  In scenarios like case A3, which features moving people and foreground occlusions, RD-VIO boasts a substantially lower APE than ARKit and ARCore. This improved performance is attributed to the newly introduced subframes strategy and the IMU-PARSAC algorithm, which effectively reduce interference from dynamic objects and bolster pose robustness in degenerate motion. In situations where the camera experiences significant occlusion (e.g., case A4), RD-VIO outshines both ARKit and ARCore. Figure~\ref{fig:ARKit-Compare} shows the trajectories generated by the mentioned algorithms and the trajectories recorded by VICON. It can be clearly observed that RD-VIO achieves more stable and robust tracking in such challenging scenarios. It is crucial to acknowledge that both ARKit and ARCore are comprehensive VI-SLAM systems. They've benefited from extensive engineering optimizations, spanning hardware, software, and chip-level enhancements. In contrast, our research centers on crafting a compact visual-inertial odometry system that strikes a balance between being lightweight and robustness.

\subsubsection{Mobile AR Application}
\begin{figure}
    \centering
    \includegraphics[width=\columnwidth]{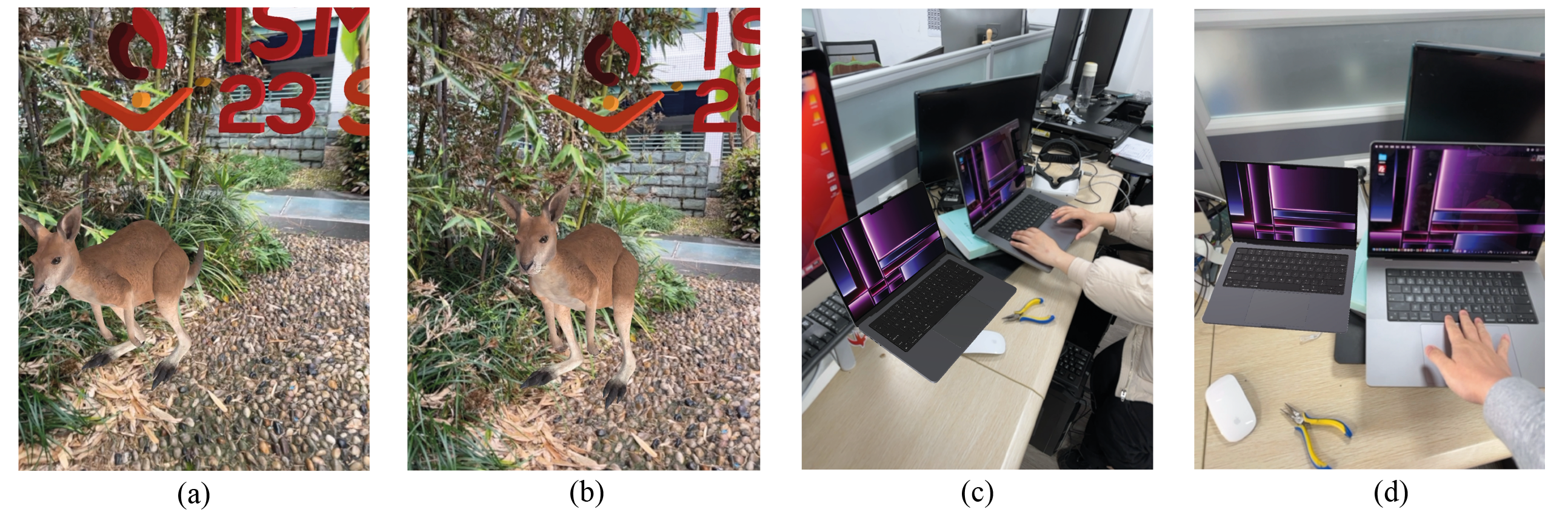}
    \caption{AR effect on a mobile phone: (a)-(b) kangaroo in the wild. (c)-(d) try the product before purchase.}
    \label{fig:fancy AR demo}
\end{figure}

We deploy RD-VIO to iOS platform and develop a simple AR demo to show its accuracy and robustness. We use 30 Hz images data with resolution $640\times480$ and IMU data which consists of  angular velocity and acceleration with 100Hz captured by the iPhone X. RD-VIO can run in real-time on mobile devices. A virtual cube and some other virtual objects are inserted into the real scenes. Figure~\ref{fig:fancy AR demo} shows two AR examples. We also compare it against with VINS-Mobile, which is one of the best open source mobile AR system. Both of them run on iPhone X. The experimental results show the superiority of RD-VIO both on pure rotation conditions and dynamic scenes~\ref{fig:teaser}. Please refer to the supplementary video for the complete results with comparisons.

\section{Conclusion}

In this paper, we propose a robust and novel VIO system which can efficiently handle dynamic scenes and pure rotational motion. By using IMU-PARSAC algorithm, dynamic feature points are removed in a two-state process. This method enables our system to effectively respond to drastic scenarios changes. For pure rotation motion, we design a subframe structure and use deferred-triangulation technique. Both of which bring us a significant improvement in degenerate motion scenes. 

We have achieved obvious better results than the baseline on EuRoc and ADVIO datasets, which proves the effectiveness of our system. Our algorithm also has a good performance on computation cost and can run on a mobile device in real-time. The AR demo on the iPhone X further demonstrates the robustness of the algorithm in challenging scenarios. It also illustrates the ability of the algorithm in the field of applications for mobile AR.

Our system still has some limitations. It could not work well when devices are in extremely challenging scenes for a long time. Especially, when there are no valid visual observations as input, our system will lose tracking inevitably. In this condition, combining some other algorithms maybe helpful, such as pure inertial odometry or wireless tracking.


%



\section*{Acknowledgments}
This work was partially supported by NSF of China (No. 61932003). The authors would like to thank Xinyang Liu for his kind help in data collection. Thanks to Danpeng Chen, Weijian Xie and Shangjin Zhai for their kind help in system fine-tuning and evaluation.



\ifCLASSOPTIONcaptionsoff
  \newpage
\fi


\bibliographystyle{references/abbrv-doi-hyperref}
\bibliography{references/bibliography}

%



%

\vspace{-15mm} 

\begin{IEEEbiography}[{\includegraphics[width=1in,height=1.25in,clip,keepaspectratio]{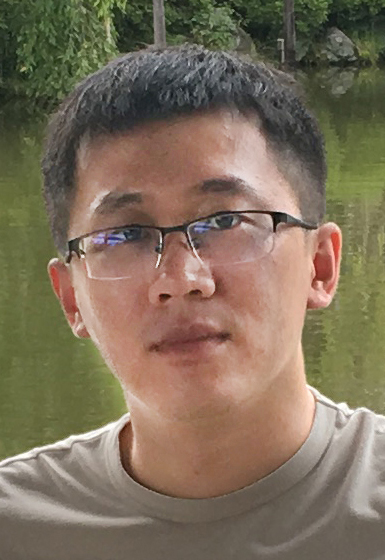}}]{Jinyu Li} received the Ph.D. degree in computer science and technology from the Zhejiang University in 2022. His research interests include 3D reconstruction, SLAM and Augmented Reality.
\end{IEEEbiography}

\vspace{-15mm} 

\begin{IEEEbiography}[{\includegraphics[width=1in,height=1.25in,clip,keepaspectratio]{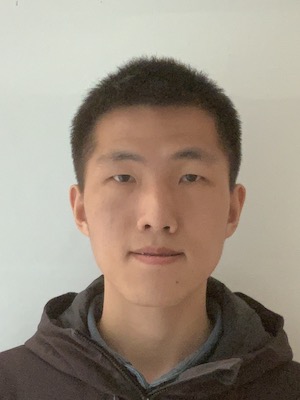}}]{Xiaokun Pan} is currently a Ph.D. student at Zhejiang University. He received the B.S. degree in Electrical Engineering from Wuhan University in 2019. His research interests include 3D reconstruction, SLAM and Augmented Reality.
\end{IEEEbiography}

\vspace{-15mm} 

\begin{IEEEbiography}[{\includegraphics[width=1in,height=1.25in,clip,keepaspectratio]{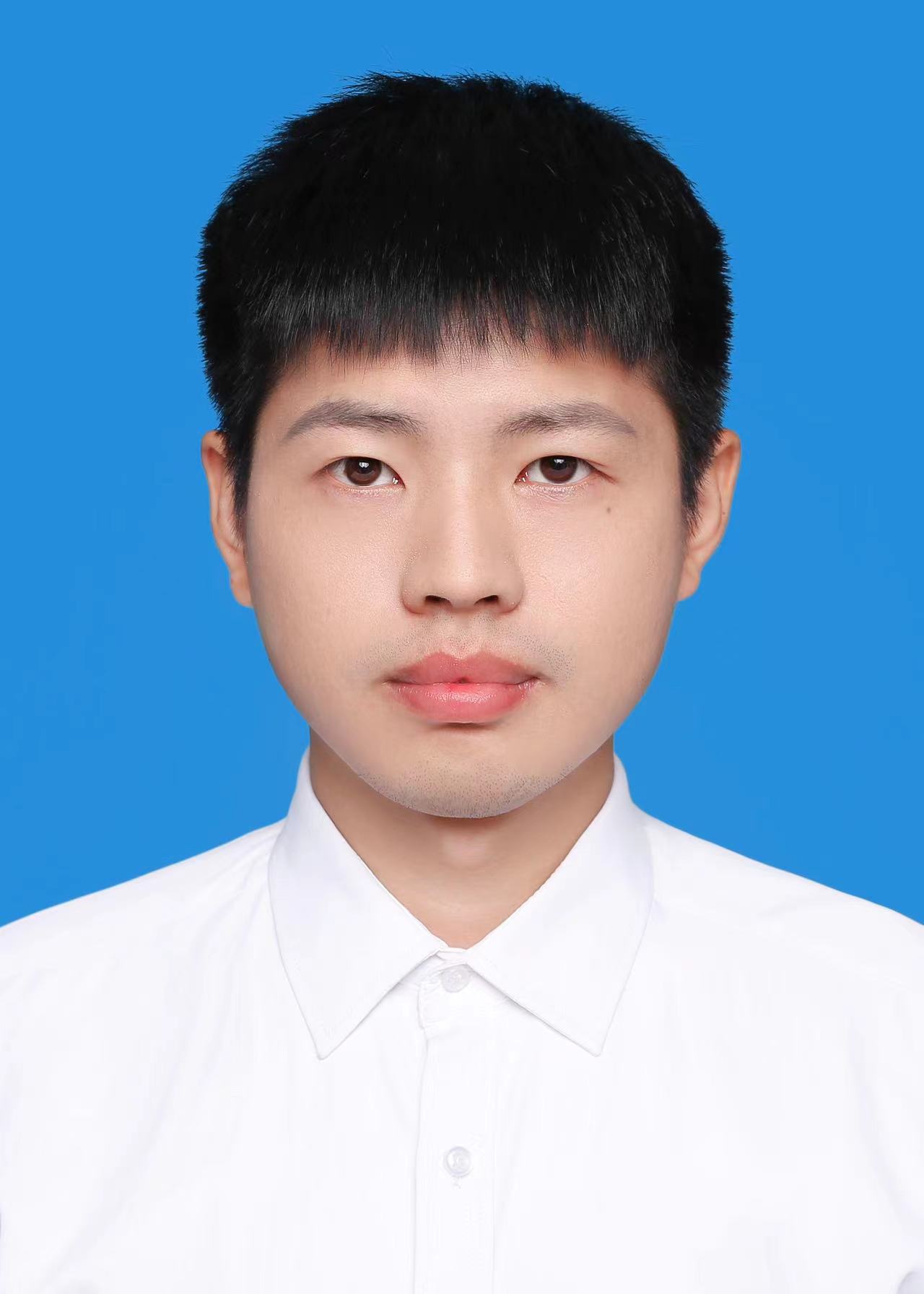}}]{Gan Huang} is curretnly a Ph.D. student at Zhejiang University. He received the B.S. and M.S. degrees in Electronic Information Engineering from China Jiliang University in 2018 and 2022. His research interests include SLAM, 3D reconstruction and visual knowledge learning.
\end{IEEEbiography}

\vspace{-15mm} 

\begin{IEEEbiography}[{\includegraphics[width=1in,height=1.25in,clip,keepaspectratio]{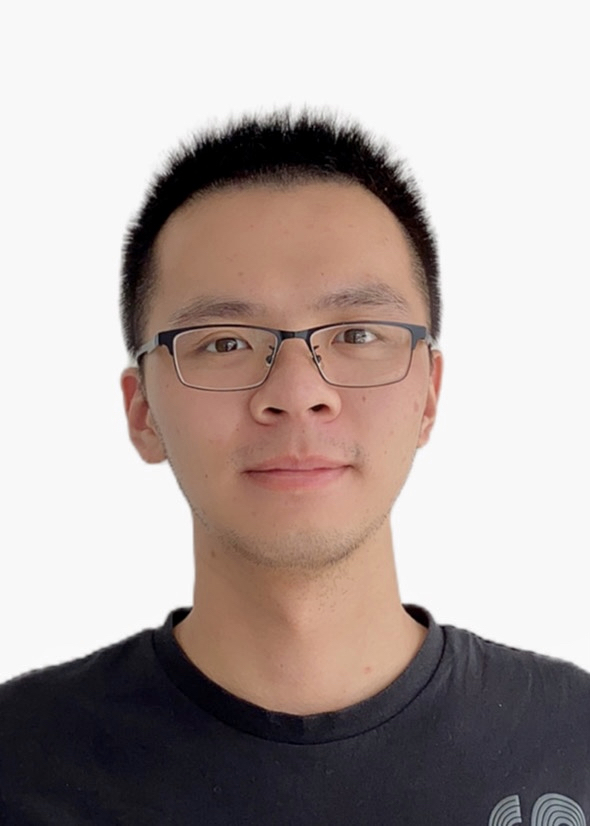}}]{Ziyang Zhang} is currently a Ph.D. student at Zhejiang University. He received the B.S. degree in Computer Science and Technology from Heilongjiang University in 2020. His research interests include HCI, SLAM, and Augmented Reality.
\end{IEEEbiography}

\vspace{-15mm} 

\begin{IEEEbiography}[{\includegraphics[width=1in,height=1.25in,clip,keepaspectratio]{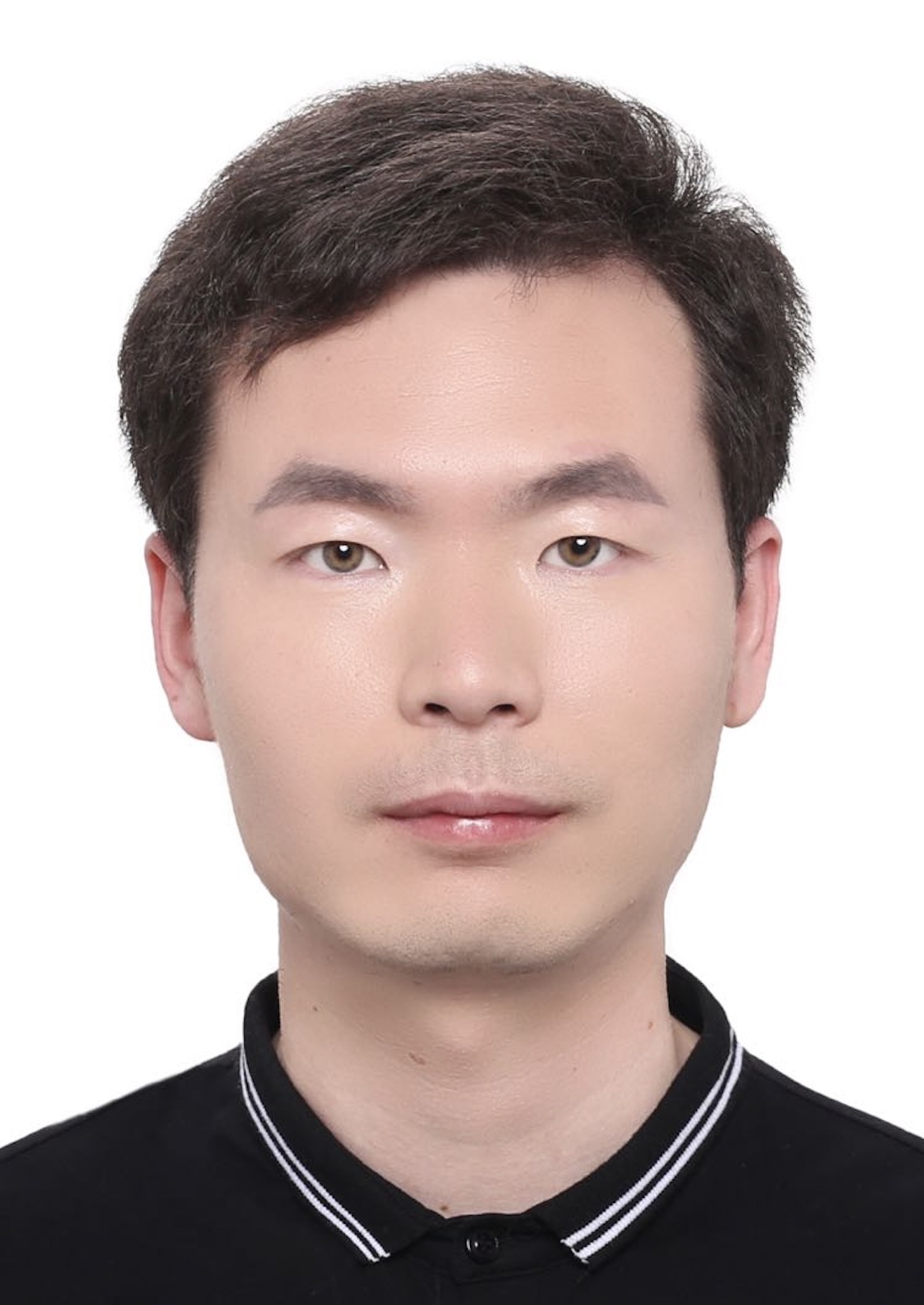}}]{Nan Wang} is currently working as Research Director at SenseTime. Before that, he was a Senior RD at Baidu Inc.. He obtained his Master’s degree in the State Key Lab of CAD\&CG, Zhejiang University in 2015, advised by Prof. Guofeng Zhang. His research interests include SLAM, 3D reconstruction, and augmented reality.
\end{IEEEbiography}

\vspace{-15mm} 

\begin{IEEEbiography}[{\includegraphics[width=1in,height=1.25in, clip,keepaspectratio]{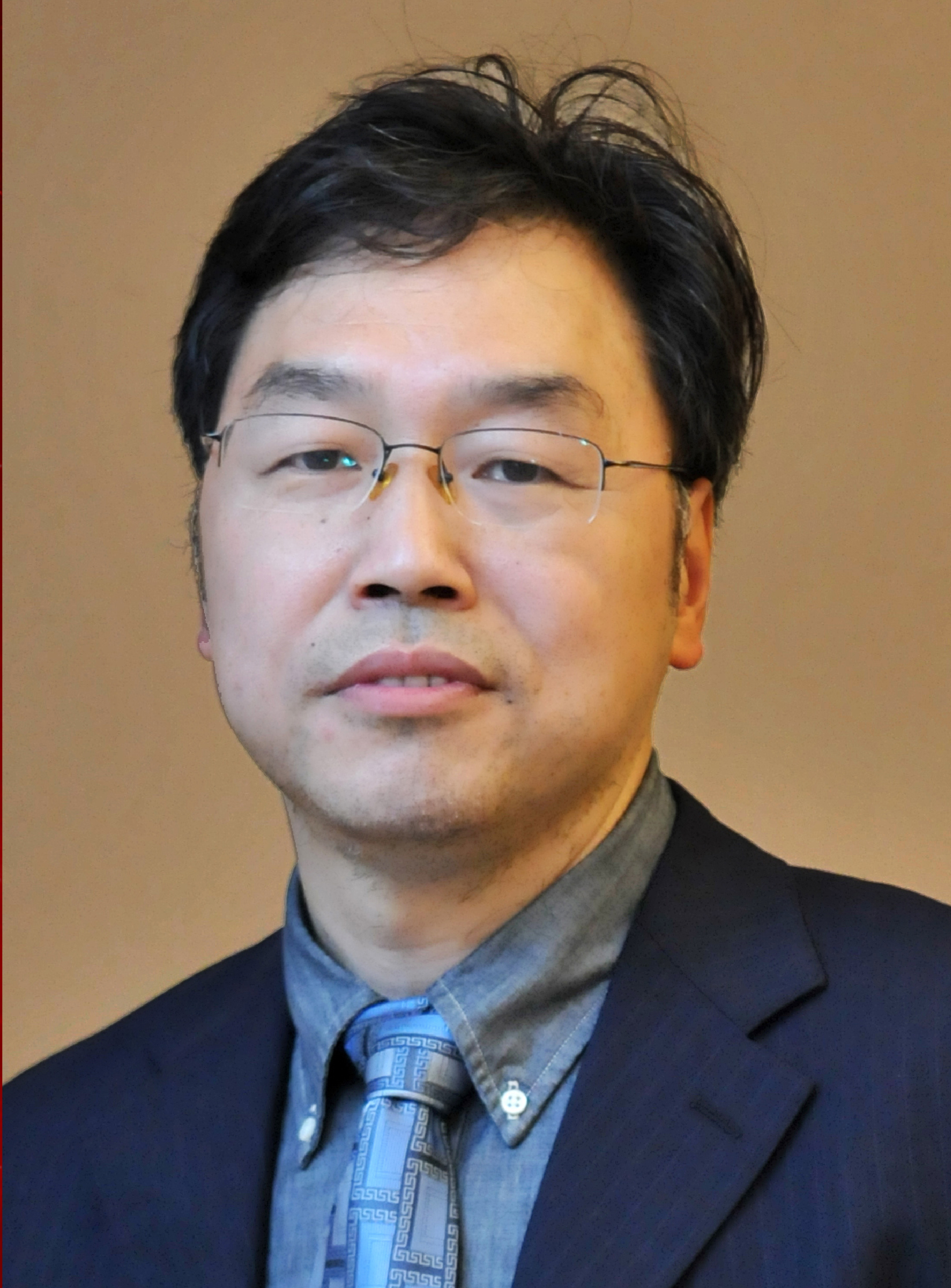}}]{Hujun Bao} is currently a professor in the Computer Science Department of Zhejiang University, and the former director of the State Key Lab of CAD\&CG. His research interests include computer graphics, computer vision and mixed reality. He leads the mixed reality group in the lab to do a wide range of research on 3D Reconstruction and Modeling, Real-time Rendering and Virtual Reality, Real-time 3D Fusion, and Augmented Reality. Some of these algorithms have been successfully integrated into the mixed reality system SenseMARS.
\end{IEEEbiography}

\vspace{-15mm} 

\begin{IEEEbiography}[{\includegraphics[width=1in,height=1.25in,clip,keepaspectratio]{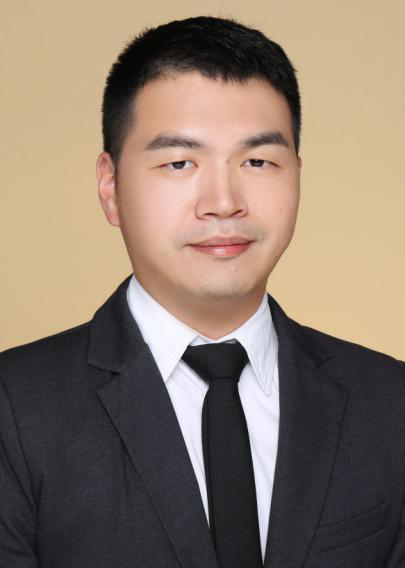}}]{Guofeng Zhang} is currently a professor at Zhejiang University. He received the B.S. and Ph.D. degrees in computer science and technology from Zhejiang University in 2003 and 2009, respectively. He received the National Excellent Doctoral Dissertation Award, the Excellent Doctoral Dissertation Award of China Computer Federation and the ISMAR 2020 Best Paper Award. His research interests include SLAM, 3D Reconstruction, and Augmented Reality.
\end{IEEEbiography}








\end{document}